%% file: main.tex
\documentclass[sigconf]{acmart}

\AtBeginDocument{%
  }

\usepackage[T1]{fontenc}
\usepackage[utf8]{inputenc}
\usepackage{microtype}
\usepackage{booktabs}
\usepackage{graphicx}
\usepackage{amsmath}
\usepackage{cleveref}
\usepackage{subfig}
\usepackage{subcaption}
\usepackage{soul}
\usepackage{cleveref}
\usepackage[]{mdframed}
\usepackage{xcolor}
\usepackage[most]{tcolorbox}
\newtcolorbox{mybox}{breakable,enhanced, colback=blue!5!white,colframe=blue!75!black}
\newtcolorbox{myboxorange}{breakable,enhanced, colback=orange!5!white,colframe=orange!75!black}
\newtcolorbox{myboxred}{breakable,enhanced, colback=red!5!white,colframe=red!75!black}

\begin{document}
\title{Kahani: Culturally-Nuanced Visual Storytelling Tool for Non-Western Cultures}

\input{content/authors}

\settopmatter{printacmref=false}  
\renewcommand\footnotetextcopyrightpermission[1]{}  
\pagestyle{empty}  


\input{content/0-abstract}

%
%

\ccsdesc[500]{Human-centered computing~Empirical studies in HCI}
\ccsdesc[500]{Human-centered computing~User studies}
\ccsdesc[500]{Human-centered computing~Empirical studies in collaborative and social computing}

\keywords{Generative AI, Culturally-nuanced Visual Storytelling, Human-AI interaction, Non-Western Culture, India}

\maketitle

\input{content/1-introduction}
\input{content/2-related-work}
\input{content/3-kahani}

\input{content/4-methodology} 
\input{content/5-findings}
\input{content/6-discussion}
\input{content/9-ethics-statement}
\input{content/7-conclusion}
\input{content/10-acknowledgements} 

\bibliographystyle{ACM-Reference-Format}
\bibliography{custom}

\appendix
\input{content/appendix/appendix}

\end{document}

%% file: content/authors.tex
\author{Hamna}
\affiliation{%
 \institution{Microsoft Corporation}
 \city{Bengaluru}
 \country{India}}
 \email{hamnaabid@gmail.com}
 \authornote{Corresponding author.}

\author{Deepthi Sudharsan}
\affiliation{%
 \institution{Microsoft Corporation}
 \city{Bengaluru}
 \country{India}}
 \email{deepthi.sudharsan@gmail.com}

\author{Agrima Seth}
\affiliation{%
 \institution{University of Michigan}
 \city{Ann Arbor}
 \country{USA}}
 \email{agrima@umich.edu}
 \authornote{Work done while at Microsoft}

\author{Ritvik Budhiraja}
\affiliation{%
 \institution{Microsoft Corporation}
 \city{Bengaluru}
 \country{India}}
 \email{ritvik19322@iiitd.ac.in}

\author{Deepika Khullar}
\affiliation{%
 \institution{Microsoft Corporation}
 \city{Noida}
 \country{India}}
 \email{deepikakhullar12@gmail.com}

\author{Vyshak Jain}
\affiliation{%
 \institution{Microsoft Corporation}
 \city{Bengaluru}
 \country{India}}
 \email{vyjain@microsoft.com}

\author{Kalika Bali}
\affiliation{%
 \institution{Microsoft Corporation}
 \city{Bengaluru}
 \country{India}}
 \email{kalikab@microsoft.com}

\author{Aditya Vashistha}
\affiliation{%
 \institution{Cornell University}
 \city{Ithaca}
 \country{USA}}
 \email{adityav@cornell.edu}

\author{Sameer Segal}
\affiliation{%
 \institution{Microsoft Corporation}
 \city{Bengaluru}
 \country{India}}
 \email{sameersegal@microsoft.com}

%% file: content/0-abstract.tex
\begin{abstract}

Large Language Models (LLMs) and Text-To-Image (T2I) models have demonstrated the ability to generate compelling text and visual stories. However, their outputs are predominantly aligned with the sensibilities of the Global North, often resulting in an outsider's gaze on other cultures. As a result, non-Western communities have to put extra effort into generating culturally specific stories. To address this challenge, we developed a visual storytelling tool called Kahani that generates culturally grounded visual stories for non-Western cultures. Our tool leverages off-the-shelf models GPT-4 Turbo and Stable Diffusion XL (SDXL). By using Chain of Thought (CoT) and T2I prompting techniques, we capture the cultural context from user's prompt and generate vivid descriptions of the characters and scene compositions. To evaluate the effectiveness of Kahani, we conducted a comparative user study with ChatGPT-4 (with DALL-E3) in which participants from different regions of India compared the cultural relevance of stories generated by the two tools. The results of the qualitative and quantitative analysis performed in the user study show that Kahani's visual stories are more culturally nuanced than those generated by ChatGPT-4. In 27 out of 36 comparisons, Kahani outperformed or was on par with ChatGPT-4, effectively capturing cultural nuances and incorporating more Culturally Specific Items (CSI), validating its ability to generate culturally grounded visual stories. 

\end{abstract}

%% file: content/1-introduction.tex
\section{Introduction}

\begin{figure*}[ht]
  \centering
  \includegraphics[width=\textwidth]{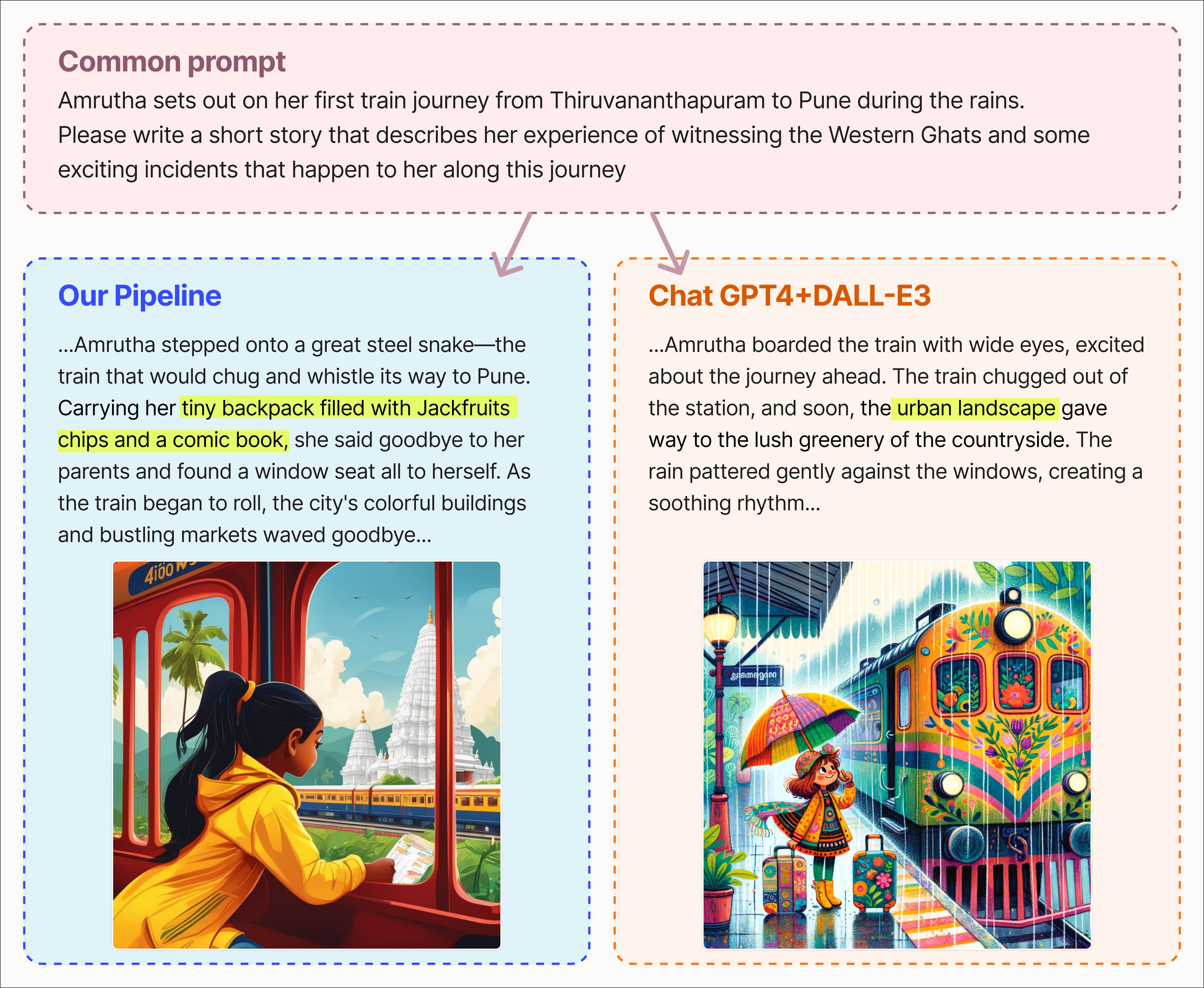}
  \caption{\textbf{Comparison of visual stories generated for a story prompt by two tools - Kahani and ChatGPT4 + DALL-E3: } Out of the two tools, we see that both visually and in text, Kahani demonstrates a significantly better ability to capture the cultural essence of the geography the story is based on.}
  \label{fig:comparison}
\end{figure*}

Visual storytelling is a powerful medium that combines visual and textual narratives to create engaging stories, making it an effective tool to convey complex ideas in a way that is both intuitive and emotionally compelling. Past works show that incorporating cultural context into these narratives enhances their relatability and resonance, making them more impactful for diverse audiences \cite{Applyculture, TranslateAds}. In education, the use of culturally familiar objects in learning content has demonstrated improved learning outcomes \cite{CulturesensitiveMaths, Hammondculturaleducation}. 
Recently, Large Language Models (LLMs) and Text-to-Image (T2I) models have been explored to generate both text and visual narratives, unlocking new possibilities for automated storytelling \cite{Yinsurvey_2024} particularly in applications such as narrative creation, screenplay writing, and content generation \cite{moviechat_song2024, synchronizedvideonarration_yang2024, mmnarrator_zhang2023}. These models have demonstrated the ability to effectively align textual and visual data, enabling the creation of coherent and contextually relevant stories \cite{he2024dreamstoryopendomainstoryvisualization, wang2023autostory}.

However, despite significant advances, past works have raised concerns about the cultural awareness of these models, including the world knowledge they encode, the ideologies their outputs mimic \cite{atari_2023}, and the extent to which they recognize diverse identities such as gender, race, and political affiliations \cite{thakur2023_genderbias, Kotek_2023, fang2024_bias, Motoki2023_politicalbias}. These concerns highlight the cultural gaps perpetuated in the outputs of these models, as they often amplify the biases and omissions inherent in their training data. 
This lack of cultural diversity often results in generated images that are unsatisfactory, inappropriate or offensive in specific cultural contexts. As the result, it exacerbates existing inequalities by reinforcing harmful stereotypes and biases \citep{struppek2022biased,qadri2023ai,cho2023dall,basu2023inspecting}. 

\begin{figure}[t]
  \centering
  \subfloat[DALL-E3]{
    \includegraphics[width=0.2\textwidth]{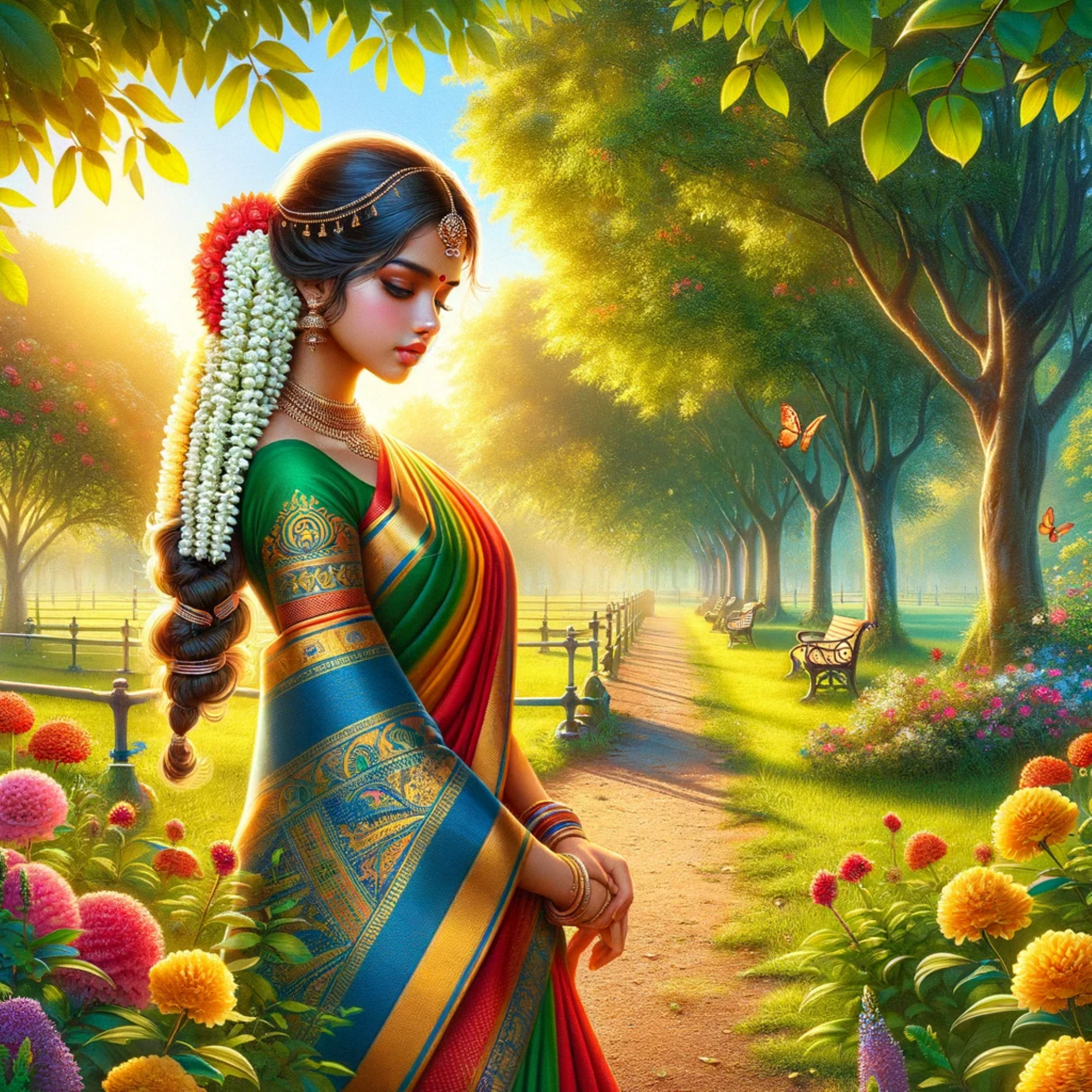}
  }
  \subfloat[Midjourney]{
    \includegraphics[width=0.2\textwidth]{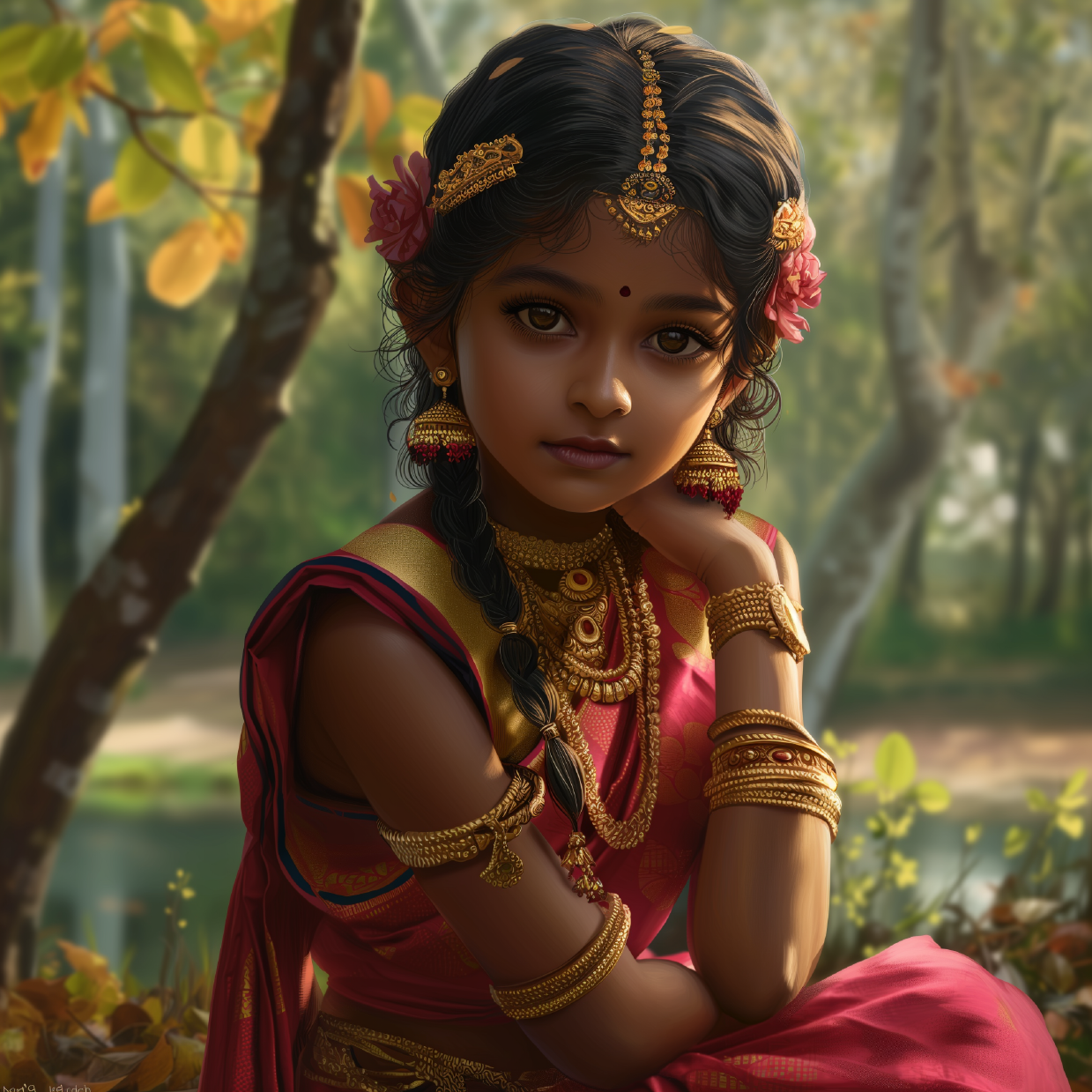}
  }
  \setlength{\belowcaptionskip}{-14.1pt}
  \caption{\textbf{Example of representational harm in state-of-the-art T2I models when asked to create image of a ``South Indian girl in a park"}.  The generated images depict ``South Indian girl" with an unreasonable
amount of ornamentation in a park, largely reinforcing the Western gaze of Indian culture.}
  \label{fig:girl}
\end{figure}

To illustrate these limitations in practice, we prompted two popular T2I models, Midjourney and Dall-E3, to create the image of a South Indian girl in a park. The generated image (Figure \ref{fig:girl}) shows a young girl with an unreasonable amount of ornamentation, largely reinforcing the Western gaze of Indian culture. This poses significant challenges to non-Western communities when generating the culturally specific outputs, as it often requires additional effort, especially in crafting prompts and translating their local or colloquial expressions into English. Addressing these gaps is crucial for developing AI systems capable of producing culturally appropriate and contextually relevant images, ensuring that the diverse experiences and identities of all users are represented. 

In this work, we introduce Kahani, a visual storytelling tool that leverages off-the-shelf LLMs (GPT-4-Turbo via OpenAI API) and Text-to-Image models (Stable Diffusion XL (SDXL) \footnote{\url{https://huggingface.co/stabilityai/stable-diffusion-xl-base-1.0}} via Automatic1111 \footnote{\url{https://shorturl.at/703Ij}} API). We utilize techniques such as the Chain of Thought (CoT) strategy \cite{10.5555/3600270.3602070} and T2I prompting methods to provide instructions and generate culturally nuanced stories and their visuals. Our work explores the following two key research questions:
\begin{itemize}
    \itemsep0em
    \item \textbf{RQ1}: How can generative AI models be guided to generate culturally nuanced stories and visuals that accurately reflect non-Western cultures?
    \item \textbf{RQ2}: What strategies can be employed to evaluate the cultural and contextual relevance of AI-generated stories and visuals?
\end{itemize}

These questions address the broader challenge of aligning generative AI systems with the expectations and experiences of diverse user communities, particularly those from non-Western cultural contexts. 

To evaluate the efficacy of our tool, we conducted a comparative user study in which the participants annotated and rated the cultural aspects of the text stories and the visuals generated by ChatGPT \footnote{\url{chat.openai.com}} and our tool, Kahani. The evaluation parameters included cultural nuances of the text and visuals, image consistency, character depiction, the use of culturally specific words, the accuracy of cultural elements in the visuals, the overall plot of the story, and the scene selection. We then conducted a mixed-methods analysis using Wilcoxon signed-rank test \cite{wilcoxonsignedranktest}, reference-based and reference-free metrics inspired by BiLingual Evaluation Understudy (BLEU) \cite{papineni-etal-2002-bleu} and Multidimensional Quality Metric (MQM) \cite{Lommel2013MultidimensionalQM} respectively to evaluate the generated cultural stories and their visuals.

Our results show that the visual stories generated by Kahani are more culturally nuanced compared to those generated by ChatGPT-4. Out of 36 comparisons (18 interviews, each with 2 stories), Kahani was better or at par with ChatGPT-4 in 27 comparisons. Kahani's visual stories were rated higher in terms of cultural relevance, accuracy of cultural elements, and overall visual appeal by participants highlighting that Kahani effectively captured cultural nuances, such as traditional clothing, local landmarks, and culturally significant activities, which were often overlooked by ChatGPT-4. Kahani's stories contained more Culturally Specific Items (CSI) \cite{Newmark2010} than ChatGPT-4 further validating its ability to generate culturally grounded visual stories.

In summary, our contributions are as follows:
\begin{itemize}
    \itemsep0em
    \item We build a novel visual storytelling tool, Kahani, that enables users in non-Western settings to create culturally appropriate stories and visual representations. Our tool is model-agnostic and can be easily adapted to future LLMs and T2I models and to different downstream tasks.
    \item We evaluated the efficacy of our tool through comparisons with visual stories generated by the state-of-the-art LLM and T2I model, ChatGPT and DALL-E3, using qualitative feedback and user-centered evaluation metrics.
    \item To facilitate future research on culturally nuanced visual story generation,we make our code implementation publicly available at GitHub\footnote{\url{https://github.com/microsoft/Kahani}}.
    
\end{itemize}

%% file: content/2-related-work.tex
\section{Related Work}

\subsection{Visual Storytelling}

Early works in story visualization have predominantly employed GAN or VAE-based approaches \cite{chen2022charactercentricstoryvisualization,li2022wordlevelfinegrainedstoryvisualization,li2019_storygan,maharana2021improvinggenerationevaluationvisual, Song2020characterpreserve}. For example, StoryGAN \cite{li2019_storygan} pioneered the story visualization task by proposing a GAN-based framework that considers both the full story and the current sentence for coherent image generation. Other early methods adopted a dataset-driven approach, relying on models trained on popular datasets such as PororoSV\cite{li2019_storygan}, FlintstonesSV \cite{maharanamohit2021integratingvisuospatiallinguistic}, or the StorySalon dataset \cite{liustorysalon2024}.  However, these methods face significant limitations due to the size and quality of existing datasets, making it difficult to generalize to diverse characters, scenes, and cultural contexts. The development of Large Language Models (LLMs) and Text-to-Image (T2I) generative models\cite{ ramesh2021zeroshottexttoimagegeneration, rombach2022_highresolutionimagesynthesis, dhariwal2021diffusionmodelsbeatgans, saharia2022paletteimagetoimagediffusionmodels} has revolutionized generalizable story visualization, achieving impressive image quality. Recent works have explored generating story visuals with diverse characters \cite{gong2023talecrafter, jeong2023zero, liustorysalon2024} and also on maintaining image consistency using either few-shot fine-tuning methods \cite{ruiz2023dreamboothfinetuningtexttoimage}, ControlNet-based methods \cite{zhang2023adding,wang2023autostory,gong2023talecrafter}, or shared attention and correspondence maps \cite{tewel2024training}. While these techniques offer better results, they are resource intensive, requiring significant computational power and technical expertise. Moreover, these studies often overlook the accuracy of cultural representation, an essential aspect of creating authentic and contextually relevant story visuals.

\subsection{Generative AI and Culture}

Large Language Models have emerged as powerful tools in Natural Language Processing (NLP), demonstrating remarkable capabilities in tasks such as text generation language translation, summarization, and complex reasoning \cite{fan2023_bibliometricreviewllm, naveed2024_overviewllm}. These models, trained on vast amounts of textual data, have revolutionized language-based AI applications \cite{fan2023_bibliometricreviewllm}. However, a critical issue is that majority of LLMs are primarily trained in English language data, introducing several limitations and biases \cite{lai2024_llmsbeyondenglish}. Extensive research has documented the harmful societal biases embedded in AI models across various demographic dimensions, including gender, race, religion and their intersections \cite{basu2023inspecting, Kotek_2023, fang2024_bias}. These biases can cause a range of real-world harms, both representational and otherwise, as highlighted by recent studies \cite{NEURIPS2021_Kirk}. For instance, Qadri et al. \cite{qadri2023ai} demonstrated that text-to-image models often fail to generate culturally accurate artifacts,instead reinforcing dominant cultural defaults and perpetuating stereotypes when depicting non-Western cultures. Similarly, Naous et al. \cite{naous2024_havingbeerprayer} found that even LLMs trained in non-Western languages, such as Arabic, exhibit Western biases and Dhruv et. al \cite{dhruv2025_aisuggestionswriting} found that Western-centric AI models homogenize writing toward Western norms, diminishing nuances that differentiate cultural expression. Together, these findings reveal the pervasive harms caused by culturally misaligned AI models, including cultural hegemony, erasure, and stereotyping \citep{struppek2022biased,cho2023dall} underscoring the urgent need to address these biases \cite{prabhakaran2022culturalincongruenciesartificialintelligence}. 
To evaluate societal biases in AI models, past work has employed methods such as  participatory research \cite{qadri2023ai}, probing \cite{saravanan2023exploring}, and prompting \cite{jha2024beyond,seth2024dosa}. \citet{bird2023typology} performs a comprehensive survey of text-to-image generative models and their associated harms. In line with prior works, they show that one of the most common harms associated with text-to-image models is the representational harm. While these works focus on eliciting the representational harms in these models, they do not necessarily propose a way to mitigate them. In this work, we complement the existing scholarship on biases in text-to-image models by developing a tool aimed at improving the representation of non-Western cultures in visual storytelling.

%% file: content/3-kahani.tex
\section{Kahani}
\label{sec:Kahani}

\begin{figure*}[ht]
  \centering
    \includegraphics[width=\textwidth]{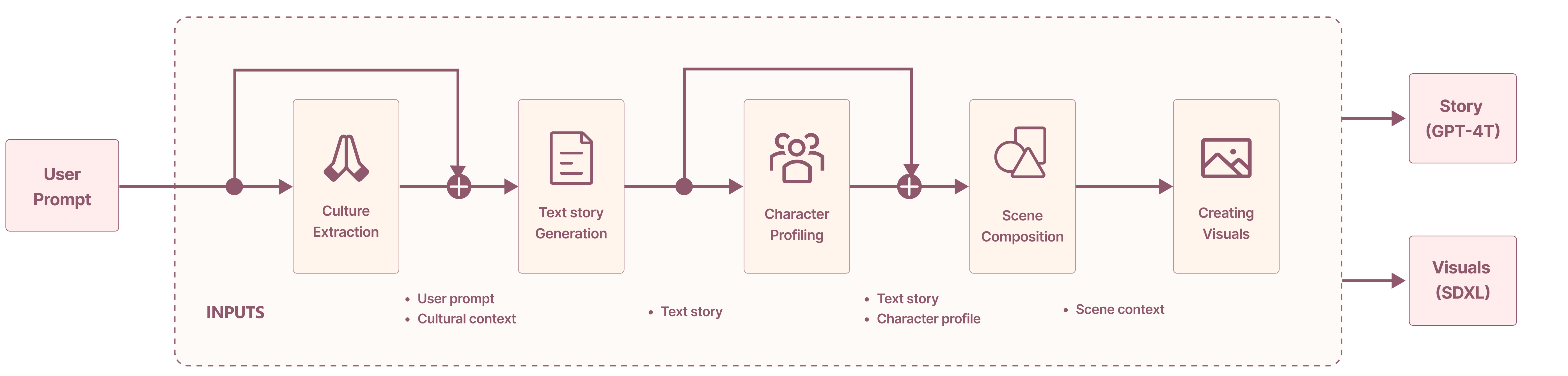}
  \caption{Overview of our Kahani Visual Storytelling Pipeline. (1) We first extract Cultural Specific Items (CSIs) from an input user prompt and expand on these CSIs; (2) We then generate the text story using the input user prompt and the expanded cultural details to add more cultural context to the story; (3) From the generated story, we extract and create character profiles consisting of visual descriptions and attributes of the characters involved in the story; (4) Using a four-act story arc as a reference, the story is segmented into distinct scenes and each scene is planned and outlined to prepare for visual generation; (5) Finally, we provide T2I prompt template and instructions to the LLM and generate T2I prompts for each scene and generate the story visuals; Stable Diffusion XL (SDXL) was used to generate the visuals, whereas GPT-4 Turbo was used for the rest of the text generations.}
  \label{fig:tool}
\end{figure*}

Kahani is a visual storytelling tool that enables users to generate culturally-grounded visual stories by using simple story prompts. It leverages OpenAI's GPT-4 Turbo API for generating cultural story texts and crafting Text-to-Image (T2I) prompts for the visuals and Stable Diffusion XL (SDXL) model to generate the story visuals. Overall, Kahani (Figure \ref{fig:tool}) consists of five primary steps: culture extraction, story generation, character profiling, scene composition, and visual generation.

 We leverage Chain of Thought (CoT) and role-play prompting techniques \cite{wu2023largelanguagemodelsdiverse} to craft the system prompts for each of the steps. To generate visuals for each scene, we use the SDXL model with GPT-4-generated T2I prompts as input. Our tool was developed in such a way that it is model-agnostic, allowing for easy upgrade of models as and when they become available. 
This section provides a brief overview of the steps in our visual storytelling pipeline.  To illustrate each step, we use the user story prompt provided below and present its intermediate outputs in the respective sections for better understanding of each step in the pipeline. The prompt templates for each step of the pipeline are provided in Appendix \ref{appendix:a-prompts} and the final visual story output for the example story prompt is provided in Appendix \ref{appendix:a-preeti-story-kahani}.

\begin{mybox}
\label{preeti-prompt}
\textbf{User prompt for the sample story (Preeti):}\\
It is a sunny morning in Dalhousie. \textbf{Preeti} and her friends are strolling the local markets. While strolling, they stumble upon a forgotten, dilapidated building with a hidden entrance. Write a story about their journey as they explore the abandoned structure, encountering eerie mysteries and thrilling encounters that test their courage and resolve.
\end{mybox}

\subsection{Culture Extraction} 

\begin{myboxorange}
    \textbf{Intermediate Output:}
    \begin{itemize}
        \item \textbf{Dalhousie:} a hill station in Himachal Pradesh, India; Indian characters are likely.
        \item \textbf{Preeti:} a common Indian female name, possibly Hindu.
        \item \textbf{Local markets:} a setting where bargaining, small family businesses, and street food are common.
        \item \textbf{Forgotten, dilapidated building:} might evoke local legends or superstitions common in Indian culture.
        \item \textbf{Exploring abandoned structures:} could involve implicit cultural risks or taboos in India.
    \end{itemize}
\end{myboxorange}

As the first step, to gather additional cultural context for the story generation, we designed a prompt that extracts all plausible cultural contexts from the user’s input story prompt based on a specified list of cultural proxies \cite{hall1976beyond, adilazuarda2024measuringmodelingculturellms} such as location, gender, religious background, etc. inspired by Newmark's Cultural Specific Items (CSIs) categories \cite{Newmark2010}. Then, we instruct the model to expand these extracted cultural contexts to provide additional information. This includes expanding transliterated words or phrases mentioned in the user prompt for better understanding. 

\subsection{Text Story Generation}

\begin{myboxorange}
    \textbf{Story generated:}
    \\\\
     In the breezy hills of Dalhousie, under a sky as clear as a sparkling stream, Preeti pulled her scarf tighter around her neck. The chilly air kissed her cheeks as she and her friends, Ankit and Isha, wandered through the bustling market...
\end{myboxorange}

Once the cultural elements are extracted, we provide them, along with the input story prompt, to our story generation prompt for incorporating additional cultural context while generating the story. We limit the length of the story to 500 words. The complete text story generated during this step for the sample story is provided in Appendix \ref{appendix:a-preeti-story-kahani}. 

\subsection{Character Profiling}

\begin{myboxorange}
    \textbf{Intermediate Output:}
      \begin{itemize}
        \item \textbf{Preeti:} A 10-year-old girl from Dalhousie with light brown skin and shoulder-length, straight dark hair. Preeti often sports a red woolen sweater over a denim dress paired with white leggings, which is suitable for the cool climate. She completes her look with sturdy brown boots and a mischievous glint in her almond-shaped eyes.
        \item \textbf{Arjun:} A boy of 11 from Dalhousie with tan skin and short, curly brown hair...
    \end{itemize}
\end{myboxorange}

In the third step, we extract the characters critical to the story arc and expand their important characteristics from the story generated in the previous step. A list of instructions was provided to the model to avoid generic, non-informative descriptions. These instructions include incorporating a visual
description of the character, geographical information and cultural characteristics of the characters based on the location and story plot.

\subsection{Scene Composition}

\begin{myboxorange}
    \textbf{Intermediate Output:}
          \begin{itemize}
        \item \textbf{Narration:} Preeti walks with the excitement of a leader on a quest. Her eyes shine like stars as she steers her gang through the colorful market. It's a sunny game of peek-a-boo with the mountains, and the street's alive with the song of Dalhousie. Each shop was a paintbox of surprises, each sniffing a tale of flavors. Friends follow, wrapped in the market's magic swirl.    
        \item \textbf{Backdrop:} "The setting is the lively market streets of Dalhousie, with the Dhauladhar mountain range looming the background. The streets are narrow and lined with small shops boasting an assortment of vibrant goods - textiles, local handicrafts, and perhaps a dhaba serving up steaming momos. Cultural accuracy is attended to with signboards in both Hindi and English, and the occasional sight of local attire like the Himachali cap. The sun plays mischievously with the scene, casting a warm glow and creating a bustling atmosphere."
        \item \textbf{Context:} "Scene 1: Preeti, with a sparkle of adventure in her eyes, is leading her group of friends confidently through the bustling market streets of Dalhousie, amidst the vibrant chaos of shop displays, street food, and local chatter."    
        \item \textbf{Characters:} \\
            \begin{itemize}
            \item \textbf{Preeti:} "Stride forward confidently, head held high, one hand slightly extended as if beckoning her friends to follow, \textless Face lit with enthusiasm and a trace of daring\textgreater."
            \item \textbf{Arjun:} "A few steps behind, looking at Preeti with admiration, \textless Smiling, with eyes full of trust and slight awe towards Preeti\textgreater."
        \end{itemize}
    \end{itemize}
\end{myboxorange}

The next step is to compose the story into scenes and plan these scenes and their setup to generate the visuals.

\subsubsection{Story Scene Segmentation}

To segment the generated stories into distinct scenes, we instruct the model to follow the story arc outlined below, ensuring that the main elements of the text story are effectively captured.

\begin{itemize}
    \itemsep0em
    \item Scene 1: Introduction of the protagonist in his context.
    \item Scene 2: Introduction of the conflict.
    \item Scene 3: Climax of the story.
    \item Scene 4: Conclusion on an uplifting note.
\end{itemize}

\subsubsection{Scene Planning}

Once the story has been divided into four scenes, we outline and plan each scene (similar to storyboarding) to prepare for visual generation. The scene planning prompt includes instructions to ensure geographical and cultural accuracy, incorporate Cultural Specific Items (CSIs), and highlight key visual details, such as background elements, facial expressions, and character poses in each scene.

\subsection{Creating visuals}

This subsection outlines the final step of the visual storytelling tool, which involves creating visuals for each of the generated scenes.

\subsubsection{Crafting T2I prompts}

\begin{myboxorange}
    \textbf{Intermediate Output:}
    Girl and Boy, ((Girl 10 years old, wearing red woolen sweater over denim dress with white leggings, confident stride, hand extended, alight with enthusiasm and daring:1.2)), ((Boy 11 years old, tan skin, navy blue jacket over yellow t-shirt, khaki cargo pants, looking on with admiring smile, intrigued eyes:1.2)), (Dalhousie market streets bustling with vibrant goods and steaming food stalls, Dhauladhar mountains in the background, narrow lanes, Himachali caps, signboards in Hindi and English, warm sunlight casting a cheerful glow), (Kids illustration, Pixar style:1.2), masterpiece, sharp focus, highly detailed, cartoon
\end{myboxorange}

To craft the T2I prompts for SDXL, we provide the GPT-4 model with the generated character profiles and generated scene details from the previous steps along with instructions on how to craft the T2I prompts. The instructions ensure that the model captures actions in scenes, spatial relationship between characters, replace local words with their visual descriptions, replace character names with their demographic identifiers etc. while generating the T2I prompts. 

\subsubsection{Visual Generation}
\label{sec: visual generation}

For the generation of the scene visuals, the generated T2I prompt for each scene is provided as input to the SDXL model (accessed via Automatic1111 WebUI API) along with a carefully constructed negative prompt to ensure the generation of safe and appropriate content. The negative prompt template is provided below.

\begin{myboxred}

\label{negative_prompts}
NEGATIVE PROMPT ="EasyNegative, blurry, (bad prompt:0.8), (artist name, signature, watermark:1.4), (ugly:1.2), (worst quality, poor detail:1.4), (deformed iris, deformed pupils, semi-realistic, CGI, 3d, render, sketch, drawing, anime:1.4), text, cropped, out of frame, worst quality, low quality, jpeg artifacts, ugly, duplicate, morbid, mutilated, extra fingers, mutated hands, poorly drawn hands, poorly drawn face, mutation, deformed, blurry, dehydrated, bad anatomy, bad proportions, extra limbs, cloned face, disfigured, gross proportions, malformed limbs, missing arms, missing legs, extra arms, extra legs, fused fingers, too many fingers, long neck, lowres, error, worst quality, low quality, out of frame, username, NSFW"
\end{myboxred}

\begin{figure}[h]
  \centering
  \includegraphics[width=\linewidth]{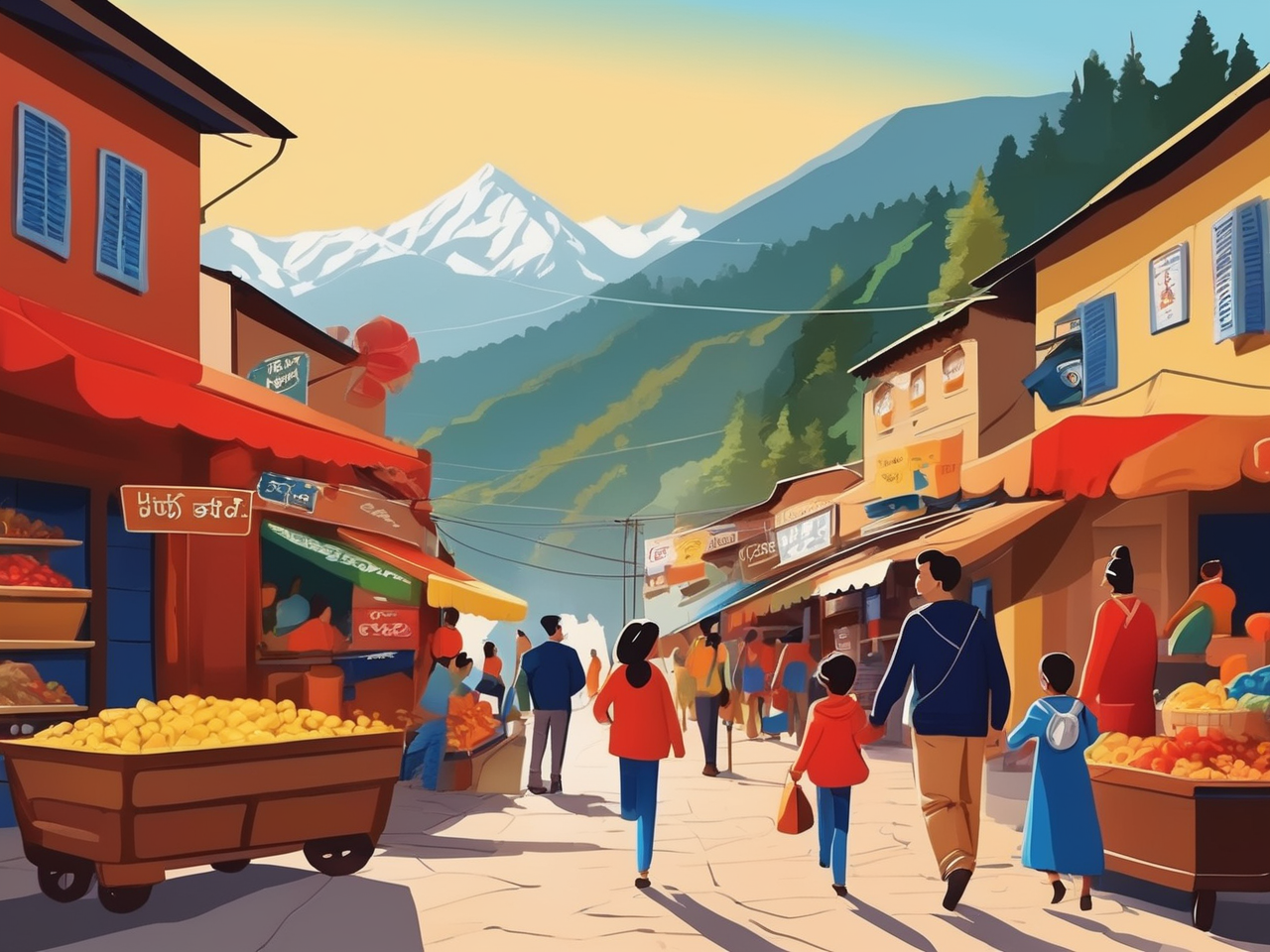}
  \caption{Scene 1 visual generated for the example (Preeti) story}
  \label{fig:Visual for Scene 1}
\end{figure}

We used the SDXL base model for text-to-image generation, followed by up-sampling the generated images using the SDXL refiner model with denoising strength of 0.5. Both generations were run for 50 steps using the DPM++ 3M SDE Karras sampler. Figure \ref{fig:Visual for Scene 1} illustrates the image generated for Scene 1 of the example visual story.

%% file: content/4-methodology.tex
\section{Methodology}

\subsection{Recruitment}

The recruitment process for our user study focused on selecting parents and teachers of children aged 4-12, as they are likely to benefit from tools that provide accessible and engaging story content. Our culturally nuanced storytelling tool is particularly suited for educational contexts, making children’s stories an ideal starting point for evaluation. Participants were recruited through a convenience sampling method, using social media platforms such as WhatsApp and Twitter. The call for volunteers included a brief description of the study's objectives which was to evaluate the extent to which the output visual stories successfully encapsulate cultural nuances.  To ensure cultural diversity, participants were asked to self-identify the state in India that best represents their cultural identity. Interested individuals were then directed to a Microsoft form to provide further information and consent to participate. 

Among the 41 responses received, we selected 25 participants through random sampling for the main study. However, only 18 participants attended. Each participant was compensated with an Amazon voucher worth 500 INR. This compensation aimed to recognize the time and effort invested by participants, aligning with ethical considerations, and fostering a sense of appreciation for their invaluable contributions to the research study.
In addition, 3 participants were selected for a pilot study conducted before the main study. The pilot study was used to assess the course and direction of the research by simulating the main study. Feedback and recommendations from participants were incorporated to streamline and refine the main study.

\subsection{Study Design}
\label{sec: Study Design}

For this study, five distinct stories and their visuals were generated using both ChatGPT4 with DALL-E3 and Kahani
using the same prompts for each story for both tools (Appendix \ref{appendix:a-prompts}). We added additional custom instructions to ChatGPT4 with DALL-E3 as shown below.

\begin{mybox}
\label{Chatgpt4-custom-instructions}
   \textbf{ Custom Instruction for ChatGPT4}: When generating a story, ensure the reading level cannot exceed grade 5 and a maximum of 500 words. For every story, please extract 4 key scenes from the story and generate visual images for these 4 scenes.
\end{mybox}

 ChatGPT4 with DALL-E3 and our tool were given the same initial prompt to generate the stories that were used for the user study:

\begin{myboxorange}
\label{userstory_prompts}
\begin{enumerate}
    \item \textbf{Nandu} stays near the farms in Alappuzha and often goes there to play with his friends and help his parents. Write a story where he and his friends are playing on the farms and come across 500 rupees dropped on the ground. While initially, they planned to keep the money to themselves, some events made them reconsider their decision, and they set out to do the right thing.
    
    \item It is a sunny morning in Dalhousie. \textbf{Preeti} and her friends are strolling the local markets. While strolling, they stumble upon a forgotten, dilapidated building with a hidden entrance. Write a story about their journey as they explore the abandoned structure, encountering eerie mysteries and thrilling encounters that test their courage and resolve.

    \item \textbf{Ojas and Omkar} are twins sent to Doon boarding school. Please write a short story about an incident that teaches them how the two adapt themselves and some important lessons they learn.

    \item \textbf{Amrutha} sets out on her first train journey from Thiruvananthapuram to Pune during the rains. Please write a short story that describes her experience of witnessing the Western Ghats and some exciting incidents that happen to her along this journey.

    \item \textbf{Bala} stays in the coastal area of Chennai. He and his friends usually play along the beach in the morning. On one such morning, they encounter an injured seagull. The initial reaction of Bala and his friends was to run away, but then they remembered an important lesson taught in their school, which led them to change their decision. Write a story along the above plot line.
    
\end{enumerate}
\end{myboxorange}

 The study was carried out as a one-hour interview over Microsoft Teams and involved three parts. 

 \begin{itemize}
    \item \textbf{Part I :}  In the first part, participants were first introduced to the basics of using Figjam, a collaborative white board environment where the story and its visuals for the study were displayed. Next, they were presented with a brief definition of culture from Newmark’s CSI categorization \cite{Newmark2010} and asked to annotate a short text paragraph and a visual scene to assess their understanding of instructions.

    \item \textbf{Part II :} After receiving instructions, participants were presented with two random pair of stories out of the ten generated along with their visuals on FigJam as a part of second part. For each story, the participants were asked to annotate the cultural aspects of the text stories and the visuals generated (Refer Figure \ref{fig:task_instructions} for the annotator instructions). To control order effects, the sequence of the stories provided was randomized and to evaluate the storytelling capabilities of the models without bias, the participants were not informed which tool was used to generate which story. During the study, we conducted a semi-structured interview where participants shared their stories preferences and experiences using AI tools for story or image generation. Interview questions can be found in Figure \ref{fig:interview questions}. 
  
    \item \textbf{Part III :}  In the last part, we supplemented our interviews with a short survey post-annotation, where the participants were asked to rate the stories and visuals on the following criteria: 
    \begin{itemize}
        \itemsep0em
        \item Text : Cultural Nuance, Culture-Specific Words, Plot of the Story
        \item Visuals : Scene Selection, Image consistency, Character Depiction, Accuracy of Cultural Elements
    \end{itemize}

    For each criteria, the participants were asked to rate on a Likert scale from 1 (Poor) to 5 (Excellent).

\end{itemize}
 
 \subsection{Data Analysis}

Post the user study, we performed both qualitative and quantitative analysis to evaluate the cultural visual storytelling capabilities of both tools for non-Western cultures.

\subsubsection{Qualitative Analysis}
We conducted a thematic analysis of the user annotations in FigJam \footnote{\url{https://www.figma.com/figjam/}} and the semi-structured interview transcripts. The process involved three key steps. First, we familiarize ourselves with the data by thoroughly reviewing the annotations and transcripts. Next, we identified multiple initial codes based on recurring phrases and ideas. Finally, we refined the coding by grouping similar codes and categorizing them into four broader themes that captured the key patterns and insights. The four identified themes were (1) Cultural relevance in the text and visuals, (2) Character depiction, (3) Story narration, and (4) Scene composition.The findings from the qualitative analysis are discussed in Section \ref{sec: Qualitative Findings}.

\subsubsection{Quantitative Analysis}

Our quantitative analysis comprises of four approaches: 
Wilcoxon signed-rank test and composite score as a part of the statistical analysis, reference-based evaluation inspired by BLEU metric \cite{papineni-etal-2002-bleu}, and reference-free evaluation inspired by Multidimensional Quality Metrics (MQM) \cite{Lommel2013MultidimensionalQM}. The findings from the quantitative analysis are discussed in Section \ref{sec:quantitative-findings}.

\begin{enumerate}
    \item \textbf{Statistical Analysis (Wilcoxon signed-rank test):}
   In this statistical analysis, we performed the Wilcoxon signed-rank test  \cite{wilcoxonsignedranktest} on the Likert scale ratings collected from participants on various criteria for all stories they were presented with in the post-annotation survey (Refer Section \ref{sec: Study Design} for evaluation criteria). This allows us to evaluate statistically significant differences between Kahani and ChatGPT-4 for each criteria measured on the visual stories.

   \item \textbf{Statistical Analysis (Composite Score):}
   For calculating the composite score we leverage the Likert ratings provided by the participants during the user study. The composite score (Equation \ref{equation:compositescore}) was calculated for each participant, story, and tool by multiplying the ratings for each parameter: Text (`Cultural Nuance', `Culture-Specific Words') and Visuals (`Image Consistency', `Character Depiction', `Accuracy of Cultural Elements').  These composite scores were then normalized to a value out of 100. Finally, we computed the mean normalized score across all participants for each story and tool.

   \begin{equation*}
        \label{equation:compositescore}
        \text{Score} = \left(
        \frac{\text{T}_{\text{CN}}}{5} \times
        \frac{\text{T}_{\text{CW}}}{5} \times
        \frac{\text{V}_{\text{CO}}}{5} \times
        \frac{\text{V}_{\text{CD}}}{5} \times
        \frac{\text{V}_{\text{AC}}}{5}
        \right) \times 100
        \tag{1}
    \end{equation*}

    where $T_{CN}$ is the rating for text on cultural nuance, $T_{CW}$ for text on culture-specific words, $V_{CO}$ is the rating for visuals on image consistency, $V_{CD}$ is the rating for visuals on character depiction, $V_{AC}$ is the rating for visuals for accuracy of cultural elements. All ratings are out of 5.
     
    \item \textbf{Reference based Evaluation:} To evaluate the quality of the participant CSI annotations and filter out any false positives, we draw inspiration from the BLEU score metric and its formulation (Equation \ref{equation:bleu}). Utilizing the reference highlights (Refer Table \ref{table:csi-categories-examples}) that the authors of the work (serving as expert annotators) created by using Newmark’s CSI categorization \cite{Newmark2010} and Aixela's classification \cite{Aixela+1999+52+78} as the theoretical frameworks to identify these items, we compared these with the CSI highlighted by participants during the user study. Precision scores were calculated for N-grams of highlighted terms (N = 1 to 4) by comparing the user-generated highlights to the corresponding reference highlights. The geometric mean of these precision scores was then computed. To account for over-highlighting, a penalty factor was applied for highlighting unnecessary terms. This penalty, inspired by the brevity penalty in BLEU, was multiplied with the geometric mean to obtain the final quality score. Once we obtain these reference-based scores across all the participants and stories, the mean of the scores across all participants per story, per tool was calculated. 

    \begin{equation}
    \label{equation:bleu}
    \text{Score} = \min(1, 1 - e^{\frac{|r|}{|p|}}) \cdot (\prod_{i=1}^{4} (precision_i)^{\frac{1}{4}}
    \tag{2}
    \end{equation}
    
    where $|r|$ is the length of the reference highlights, $|p|$ is length of the user highlights , Precision i is the precision of N-gram overlaps for N values from i=1 through 4. 

    \item \textbf{Reference free Evaluation:}  While the reference-based metric captures the quality of the user annotations by penalizing over-highlights and filtering out false positives, it does not account the degree of cultural relevance of the highlighted CSIs to the geography the story is based on. To assess this, inspired by Multidimensional Quality Metrics (MQM), for each CSI identified by the participants based on their understanding of Newmark's CSI categories during the user study, their feedback on its cultural relevancy was used to assign a score based on the scale provided below: 

    \begin{itemize}
        \itemsep0em
        \item 1 if the element is culturally appropriate for the states
        \item 0 if the element is common or generic across the world
        \item -1 if the element is culturally inappropriate for the given state
    \end{itemize}

    For each story, we take the average of the scores given by all the participants who annotated that story for the two tools and then take the average score across all the stories per tool. 
    
\end{enumerate}

%% file: content/5-findings.tex
\section{Findings}
\label{sec:findings}

\subsection{Qualitative Findings}
\label{sec: Qualitative Findings}

\subsubsection{Cultural relevance in text and visuals}
\label{sec: Cultural relevance}
Participants observed that, in contrast to the text produced by ChatGPT, the story generated by Kahani offered more vivid and immersive descriptions of settings and locations, creating an experience that felt grounded and authentic. Many participants appreciated how Kahani’s narratives captured sensory details such as sounds, smells, and visual cues, helping them to visualize the environment of the story more clearly. For instance, participant P14 remarked, \textit{``I liked how the story describes the local markets. The sounds of haggling, the rich smells of street food, and the colorful displays of small family shops surrounded them...''}. Such descriptions enhanced the engagement of the participants, as the stories felt closer to real-life experiences.

Participants also emphasized the cultural richness and specificity of the tool’s output, appreciating its integration of local and regionally significant elements. For example, Participant P10 noted, \textit{``The text describes Alappuzha well by including the paddy fields, the backwaters, etc.''}. Similarly, they observed that the Kahani-generated story was explicit in mentioning popular cultural references associated with the different regions. For example, P16 remarked, \textit{``...Gujarat is famous for their jalebi... Bengalis like to eat rice''}. P2 further observed that Kahani incorporated recognizable cultural references over generic ones, making the story feel more authentic and relatable. They stated, \textit{``...a cricket ball, some sweets, and a kite instead of generic toys''}. Such inclusions resonated with participants, as they reflected everyday experiences and culturally meaningful symbols.

Participants also appreciated the accuracy of the names, locations, and landmarks referred in the story, noting that they were factually aligned with the regions depicted. For instance, P14 reflected on how the story tied back to the historical context of Dalhousie, stating, \textit{``.. back to the history of Dalhousie, especially...the attached chest...''}. This accuracy not only enriched the narrative’s depth but also enhanced its credibility.

The relatability of the characters’ actions and experiences was another aspect participants valued. For example, P4 noted how the mention of bedtime storytelling felt familiar, stating,  \textit{``..commonly seen in India where grandmothers tell bedtime stories''}. Similarly P1 commented on a relatable scene,\textit{``...she giggled as she watched the monkey from her window...''}. This highlights how the participants connected with these everyday interactions. The relevance of character names was also appreciated, which created a sense of belonging and fostered a strong connection between the characters and the participants, as P2 noted \textit{``.. The names that were used Gopal, Mina, and Arun seemed very relevant, similar to common South Indian names used in Kerala...''}. 

While, participants appreciated the inclusion of words from the local language , such as P15's remark, \textit{``The mention of Karuthamma is a good local reference''}, they noted that such references might pose challenging for those unfamiliar with the culture. Participants suggested that providing additional context or explanations for certain local terms could enhance the accessibility of the stories for broader audiences.
Overall, the feedback highlights Kahani’s effectiveness in crafting narratives that do more than just tell a story—they immerse the reader in the local context through detailed sensory descriptions and culturally relevant themes, making the experience feel meaningful and authentic.

\subsubsection{Character Depiction}
\label{sec: Character Depiction}

With respect to the images generated by the two systems, participants noted that the visuals from Kahani were much more relatable and authentic. For example, P3 complimented that the skin complexion of the characters generated in Kahani's visuals mirrored how boys in India look. Similarly, P35 remarked, \textit{``The kid here is Indian...there's a familiarity in the picture''}. This sense of familiarity made the participants feel more connected to the visuals, as they closely reflected the cultural context.
The cultural authenticity was further enhanced by the depiction of traditional clothing, as noted by P12, \textit{``In the story, the elderly person is wearing the `Veshti' and the shirt, which we call it `Thorthu' draped around his shoulders''}. Participants appreciated how such clothing choices reflected everyday attire in many Indian households, further grounding the visuals in reality.

Participants also observed that the visuals that generated by Kahani captured emotional depth and valence, highlighting how the emotions in the story reflected the facial expressions of the characters. P8 commented on this, stating \textit{``excitement.., his Facial expression..well depicted''}. In addition, participants found that the activities that the characters are doing in stories generated by Kahani were much more relatable to their lived experience. As P8 remarked, \textit{``I can imagine my kid pressing her nose against the train window''}.

In contrast, participants noted that despite the fact that the text story has Indian names and locations, the visuals produced by DallE-3 in ChatGPT appeared eurocentric, depicting features such as blond hair, white skin tones and clothing that are less common in Indian contexts.

\subsubsection{Story narration}
Participants noted that apart from the plot being seen as engaging and able to hold readers' attention, Kahani generated stories that were perceived as avoiding common stereotypes. For example, P11 appreciated the portrayal of female characters, where the narrative deviated from traditional gender stereotypes and fearful roles.  \textit{``...is not making it up that as a female lead, she's scared or something which would find in other stories like fairy tale types.''}, P7 similarly appreciated the depiction of fantastical or unusual creatures in Kahani's story leaned toward a more positive and rational portrayal, avoiding the common trope of associating them with fear. P7 remarked, \textit{``...more positive in its treatment of the eerie creatures playful monkey as it reinforces scientific thought, rather than the previous one [GPT-generated], which had images of a stereotypical monster.''}. This findings concurs with past work, which has shown that the outputs from most of the current text-to-image models are stereotypical and reinforce existing biases \cite{qadri2023ai,jha2024beyond}. Participants found that Kahani’s stories were a step toward overcoming this limitation by presenting more nuanced and inclusive representations.

In addition to avoiding stereotypes, participants like P5 appreciated Kahani's reflection of religious diversity, an element they found missing in the text generated from ChatGPT.  They noted that this diversity was an important cultural aspect that helped the stories feel more representative of India’s multifaceted communities. However, they also felt that Kahani's stories sometimes included superfluous phrases, which they felt could make the text tedious or increase the reading level. When deciding which one they liked better, they emphasized that while the stories produced by ChatGPT were easier to understand, they were significantly limited in capturing cultural nuances.

\subsubsection{Scene Composition} 
Participants highlighted that Kahani effectively integrates culturally specific landmarks into scene compositions, reinforcing a strong sense of place. This was evident in comments such as\textit{``The lighthouse will definitely be there in Chennai''(P3)} and \textit{``...Padmanabhan temple in Trivandrum...''} (P12). They also appreciated the tool’s attention to architectural details, with one participant noting, \textit{``When I saw Dalhousie, I picked the version where the British history of Dalhousie, architecture, and landscape was accurately presented''(P2)}. Beyond architectural accuracy, Kahani’s backgrounds were praised for their realism, as reflected in remarks like \textit{``...rolling hills seem more like the Himalayas''(P2)} and \textit{``...that reminds me of Kerala—coconut trees, houseboat...''(P5)}.

Beyond structural elements, participants found Kahani’s color choices more natural and visually authentic. One remarked, \textit{``the color of the skin and the hair color used in the illustrations was much more relatable'' (P10)}, underscoring the importance of culturally sensitive visual representation. In contrast, while ChatGPT-generated images adhered more closely to scene descriptions, participants found them to be more generic, lacking the cultural depth and specificity that made Kahani’s outputs feel more immersive and recognizable.

\subsection{Quantitative Results}
\label{sec:quantitative-findings}
\subsubsection{Statistical analysis of text and visuals}
\label{sec:statistical-analysis-findings}

\paragraph{Wilcoxon signed-rank test}

The results of the Wilcoxon signed-rank test, summarized in Table \ref{table:parameter-analysis}, reveal statistically significant differences in four dimensions: Cultural Nuances (p=0.04), Cultural Specific Words (p=0.03), Image Consistency (p=0.01), and Accuracy of Cultural Elements (p=0.02). These findings suggest that Kahani outperformed ChatGPT in capturing cultural nuances and maintaining image consistency, as well as in the accurate representation of cultural elements and use of cultural specific words.

In contrast, no significant differences were observed in dimensions like Plot of the Story (p=0.12), Scenes Selection (p=0.55), and Character Descriptions (p=0.10) since their p-values are greater than the threshold of 0.05. This implies comparable performance between the two tools on these dimensions.

\begin{table}[!htb]
    \begin{minipage}{\linewidth}
      \centering
      \small
        \begin{tabular}{@{}llll@{}}
        \toprule
        Metric & \textbf{W} & \textbf{Z} & \textbf{p} \\ 
        \midrule
        \textbf{Cultural Nuances} & 11.00 & 89.00 & \textbf{0.04} \\
        \textbf{Cultural Specific Words} & 8.00 & 70.00 & \textbf{0.03} \\
        \textbf{Image Consistency} & 11.00 & 81.50 & \textbf{0.01} \\
        \textbf{Accuracy of cultural elements} & 13.00 & 86.00 & \textbf{0.02} \\
        Plot of the story & 10.00 & 58.50 & 0.12 \\
        Scenes selection & 24.00 & 152.00 & 0.55 \\
        Character Descriptions & 12.00 & 114.00 & 0.10 \\
        \bottomrule
        \end{tabular}
        \caption{Wilcoxon signed-rank (W), Z-score (Z), and p-value for comparing Kahani and ChatGPT across different metrics.}
        \label{table:parameter-analysis}
    \end{minipage}%
    \vspace{1em}
    \begin{minipage}{\linewidth}
      \centering
      \small
        \begin{tabular}{@{}lc|cccc@{}}
        \toprule
         &  & \multicolumn{2}{c}{Score (Mean)} & \multicolumn{2}{c}{Score (Std.)} \\
        Story & Count & Ours & ChatGPT & Ours & ChatGPT \\ 
        \midrule
        Amrutha & 7 & \textbf{30.49} & 21.01 & 41.42 & 35.93 \\
        Bala & 8 & \textbf{28.32} & 19.74 & 29.99 & 34.48 \\
        Nandu & 8 & \textbf{19.51} & 16.49 & 15.32 & 21.41 \\
        Ojas & 7 & \textbf{31.32} & 17.32 & 29.37 & 27.78 \\
        Preeti & 6 & \textbf{19.39} & 17.23 & 25.12 & 24.02 \\ \bottomrule
        \end{tabular}        
        \caption{Count of Interviews and the distribution of the scores based on the composite index}
        \label{table:score-parameter-5}
    \end{minipage} 
\end{table}

While the Wilcoxon signed-rank test identified statistically significant differences between the tools for certain metrics, it does not directly indicate which tool performed better. Hence, to address this limitation and gain a clearer understanding of tool performance, we calculated composite scores from the Likert ratings and analyzed the distribution of the scores based on the composite index. 

\paragraph{Composite Score}

Using the composite score (Equation \ref{equation:compositescore}), we find that of the 36 comparisons (18 interviews, each with 2 stories), Kahani
was better or at par with ChatGPT in 27 comparisons. Furthermore, in each interview, in at least one of the comparisons, users preferred the story and visuals generated by our tool. 

We observe from Table \ref{table:score-parameter-5} that Kahani's mean scores across all the metrics and participants are higher than those of the ChatGPT + DallE-3. This indicates that on an average, our tool performed better. For stories like Amrutha, Bala and Ojas, our tool performed significantly better whereas for stories like Nandu and Preeti, while our tool still outperformed the comparison, the margin of improvement was not as pronounced as observed in the other stories. From the given table, we also observe that the standard deviation for both the tools are large, as the responses were subjective and highly dependent on the individuals and their understanding of culture.

\subsubsection{Reference based Evaluation}

\begin{figure}[!ht]
  \centering
    \includegraphics[width=0.45\textwidth]{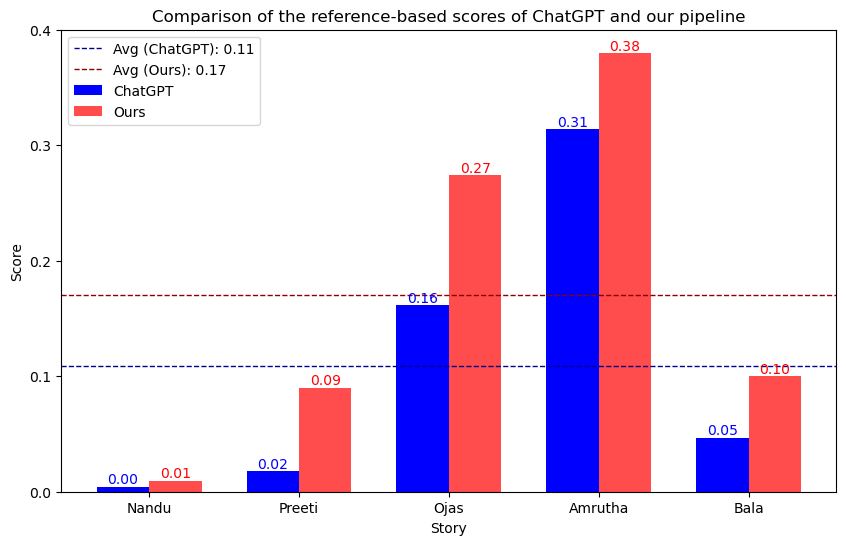}
  \caption{Comparative analysis of Reference-based metric scores}
  \label{fig:bleu}
\end{figure}

The reference based evaluation scores and visualization are shown in Table \ref{table:bleu-scores} and Figure \ref{fig:bleu} respectively. From these results, we can derive that Kahani 
was either on par or outperformed ChatGPT on average for each of the story across all the participants assigned they were assigned to. On average, across all the five stories, our tool scored 0.17 (out of 1) compared to ChatGPT with a score of 0.11. We note that Amrutha (Story 4) and Ojas (Story 3) were two of the best performing stories. Despite Kahani 
performing better, we notice how low the scores are in comparison to the maximum achievable score of this metric, which is 1. This is due to the observation that participants often highlighted long sentences rather than specific words or phrases in the generated stories. This could be attributed to the inherently ambiguous and subjective nature of defining culture. This behaviour resulted in higher penalties and poor N-gram overlap precisions in the reference-based evaluation scores. 

\begin{table}[ht]
    \begin{minipage}{\linewidth}
      \centering
        \begin{tabular}{p{3cm}cc}
        \hline
        Story & ChatGPT Score & Our Score \\
        \hline
         Nandu & 0.00 & 0.01  \\
         Preeti & 0.02 & 0.09  \\
         Ojas & 0.16 & 0.27  \\
         Amrutha & 0.31  & 0.38  \\
         Bala & 0.05 & 0.10  \\
        \hline
        \textbf{Overall Average} & \textbf{0.11} & \textbf{0.17} \\
        \hline
        \end{tabular}        \captionsetup{justification=centering}
        \caption{Reference based evaluation scores \\ (inspired by BLEU metric)}
        \label{table:bleu-scores}
    \end{minipage}%
    \vspace{1em}
    \begin{minipage}{\linewidth}
      \centering
      \large
            \begin{tabular}{@{}lcc@{}}
            \toprule
            \multicolumn{3}{c}{Avg. CSI scores}              \\ \midrule
            Story            & ChatGPT       & Ours          \\ \midrule
            Nandu            & 2.50          & 5.88          \\
            Preeti           & 0.33          & 3.00          \\
            Ojas             & 1.43          & 4.86          \\
            Amrutha          & 1.71          & 3.14          \\
            Bala             & 0.63          & 1.88          \\
            \textbf{Average} & \textbf{1.32} & \textbf{2.54} \\ \bottomrule
        \end{tabular}        \captionsetup{justification=centering}
        \caption{Reference free evaluation scores \\ (inspired by MQM)}
        \label{table:mqm-scores}
    \end{minipage} 
\end{table}

\subsubsection{Reference free Evaluation}
The reference free CSI evaluation scores are provided in Table \ref{table:mqm-scores}. From these scores, we observe that Kahani 
performed better than ChatGPT across all five stories with a score of 2.54 with Nandu (Story 1) and Ojas (Story 3) having the most number of CSIs especially those relevant to the culture the story was set in . Since this reference free evaluation metric takes into account the cultural relevancy of the CSIs highlighted and based on the severity scale provided rewards/penalizes for each highlighted text, we observed that our tool not only captured a lot more CSIs compared to ChatGPT, but they were also culturally relevant to the geography the stories were set in. These shows that Kahani was able to capture more cultural nuances compared to ChatGPT

Based on the rationale for their scores and the observations from the results, we also found the need for a more broader severity scale that would capture the essence of cultural relevance even better. Additionally, we noticed that participants could easily identify what felt wrong in the story and visuals, suggesting that cultural elements are deeply ingrained in their lives, making it difficult to capture every detail.

%% file: content/6-discussion.tex
\section{Discussion} 
We now discuss our findings and present concrete challenges in improving cultural representation in T2I models. 

\paragraph{\textbf{Aspirations, Relatability, and Diversity}}

Following the annotations, participants provided feedback on areas for system improvement. Many emphasized the importance of personal alignment in the story, where the generated visuals reflect the values they want to pass on to their children. For example, participants expressed a desire for more gender-diverse characters and scenes depicting outdoor activities and traditional games common in India, such as a group of kids playing with mud. This feedback underscores the need for a customizable system that can adapt visual storytelling to reflect users’ aspirational values and cultural preferences.

One challenge in personalizing representations to a given culture is determining how to generate relevant images without reinforcing stereotypes. For example, while cricket is one of the most popular sports in India, but how does one ensure that models do not consistently associate cricket as the default representation of sports in India while still considering it as a valid option during generation? This issue of over-reliance on cultural defaults was highlighted by a participant who remarked,\textit{``Why are the boys in the school shown playing cricket and not other sports? Not that it is wrong, but I would like to see diversity'' (P5).}
Similarly, other participants expressed the desire for greater visual diversity across multiple dimensions. They suggested increasing the representation of females in images and avoiding common associations, such as linking Indian schools primarily with subjects like Math and Science. Striking this balance between relevance and diversity will be key to ensuring that the generated visuals reflect a richer and more inclusive depiction of culture without relying on stereotypical portrayals.

While color, clothing, and other visual elements play an important role in making visuals more appealing and relatable (see Findings \ref{sec: Character Depiction}), another critical factor is the presence of text within the images. Participants gave higher ratings to models where even one visual contained text, such as a billboard that explicitly referenced a location or brand. However, participants noted that too much text or visual detail can make the image feel cluttered. Thus future work should explore how to balance text in the image, ensuring it adds meaningful context and gives a sense of relatability to the user.

\paragraph{\textbf{Halo Effect and Accuracy}}
Since the images for both ChatGPT and Kahani are generated by text-to-image models, every image generation is inherently a stochastic process. Hence, for any given prompt, the quality of the images can vary, and not all images produced by a system will be consistently good or bad. We observed a halo effect in user responses, where participants who liked the story or visuals from one system tended to rate that system higher across all metrics. This became apparent when we probed them with questions like `\textit{Could you say more about what you liked about the story?}' or `\textit{What is the other story lacking?}'. The responses often indicated a lack of specific reasoning, with participants replying:\textit{``I can not say exactly what is missing but I like the prior version''} or \textit{``It just feels better.''}

\begin{figure}[t]
  \centering
  \subfloat[Without any control]{
    \includegraphics[width=0.2\textwidth]{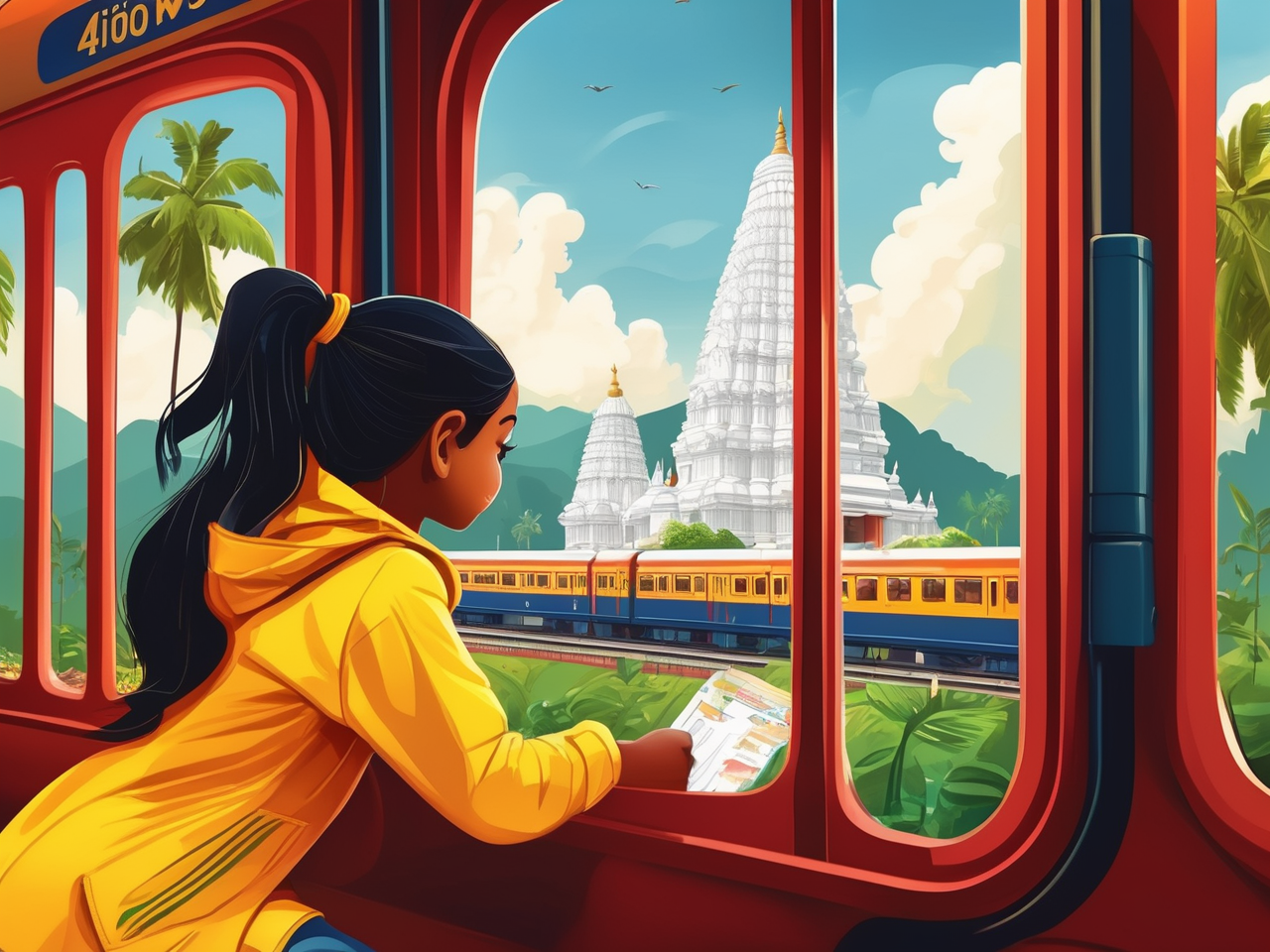}
  }
  \subfloat[With control]{
    \includegraphics[width=0.2\textwidth]{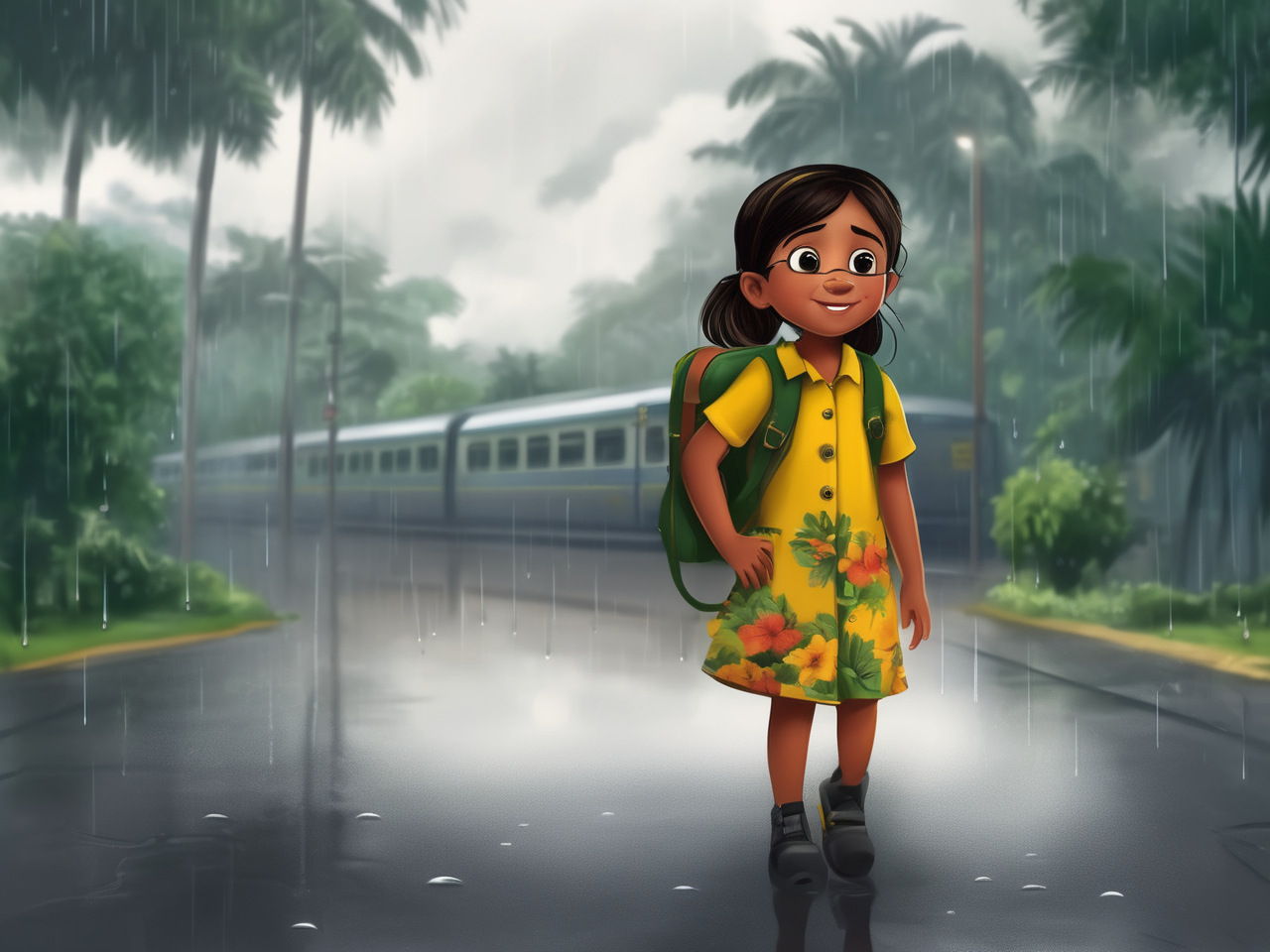}
  }
  \caption{Using ControlNet to guide did not help}
  \label{fig:controlnet-comparison}
\end{figure}

We experimented with approaches mentioned in prior works, such as TaleCrafter \cite{gong2023talecrafter}, by using ControlNet to guide the image generation process through bounding boxes and keypoints. However, we found that allowing the model to generate images freely resulted in better overall quality and more coherent scene composition. 
For example,  free generation was able to incorporate complex elements, such as a temple and a train crossing, more effectively (Figure \ref{fig:controlnet-comparison}). While in our experiments we only conditioned the placement of the main characters and not the background, we believe it was due to the complexity of the scene with multiple elements. 

While the image generated by Kahani reflected more cultural nuances, the participant observed factual inaccuracies in the visuals. For example, in a story set in the coastal area of Alappuzha, many participants questioned whether mangoes actually grew in that region. In another story, an INR 500 note was incorrectly depicted as an INR 5000 note, a denomination that does not exist. Similarly, participants pointed out a scene where a story about a low-income family showed a large heap of money, which they found inconsistent with the narrative. These errors highlight the need for methods to evaluate intermediate outputs, enabling the identification and resolution of inconsistencies during the generation process to ensure accuracy.

\paragraph{\textbf{Limitations and Future Directions}}
To address hallucinations in both text and visuals, there is a need to integrate external knowledge sources, such as Wikipedia, for fact-checking and accuracy. Expanding the system’s cultural knowledge through validated sources, including internet searches and knowledge bases, could further improve the reliability of story and visual generation. Additionally, there is a need to explore techniques beyond prompting to improve generation speed, maintain character consistency, and incorporate richer cultural details.
Drawing from our learnings about the need for personalization, it is critical that such systems allow 
users to iteratively refine stories and visuals through system feedback for enhanced user experience. A key challenge is the time required by our tool to generate images, which we are actively working to reduce.  

While our study offers important insights into cultural story generation and evaluation, it has a few limitations. To begin with, our evaluations focused on the outputs of the two system and not the process the participants used to generate those outputs. Additionally, although both Kahani and ChatGPT are multi-turn tools, our evaluation was limited to their initial outputs, missing an opportunity to assess how they handle iterative feedback or specific editing prompts—an aspect that could offer deeper insights into their responsiveness to evolving themes and cultural nuances. Our comparative analysis was also restricted to ChatGPT; expanding the study to include other popular tools would provide a broader understanding of how different models navigate cultural knowledge. While our work prioritized incorporating cultural nuances in images, character consistency in visual storytelling received less emphasis, presenting another area for future exploration. Finally, more user studies are needed with a broader and more diverse participant pool, covering a wider range of regions and cultural contexts. Casting a wider net to gather more perspectives would help us better understand the tool’s limitations and guide improvements in its overall performance.

%% file: content/9-ethics-statement.tex
\section{Ethics Statement}

\label{sec:ethics}

We use the framework by \citet{10.1162/tacl_a_00041} to discuss the ethical considerations for our
work.

\paragraph{\textbf{Institutional Review Board (IRB)}} 
All aspects of this research were reviewed and approved by the Institutional Review Board of our organization. 

\paragraph{\textbf{Annotators}}
The identity of the participants for the user study were anonymous to us and no personal information was collected. The participants were compensated with an Amazon voucher worth 500 INR for their participation. Participants were offered a pressure-free, safe environment throughout the study. They were allowed to skip any interview question if they felt uncomfortable or uneasy with the generated story or image. Additionally, participants had the option to opt out from the study at any time if the generated content was unintentionally perceived as uncomfortable, inappropriate, or offensive. The characters in the visuals and stories generated are fictional and any similarity to actual people living or dead is purely coincidental. 

\paragraph{\textbf{Annotation Guidelines}}
Annotation guidelines were provided to all participants at the beginning of the user study. Our prompts were carefully constructed and reviewed with appropriate negative prompting, to ensure the generation of safe and appropriate content (Refer Section \ref{sec: visual generation}).

\paragraph{\textbf{Compute/AI Resources}}
All our T2I experiments using SDXL (Automatic1111 Web UI) were setup and run locally on one A100 80 GB PCIE GPU. The API calls to the GPT models were done through the OpenAI service for Kahani and for the comparative ChatGPT with DALL-E3 analysis, we used the ChatGPT web interface. 

%% file: content/7-conclusion.tex
\section{Conclusion}

In this paper, we designed and implemented a culturally grounded visual storytelling tool called Kahani that focuses on improving the representation of non-Western cultures. Our user study showed that participants preferred outputs generated by Kahani to the state-of-the-art ChatGPT-4 with DALL-E3, making it a good first attempt to help creators in the Global South create culturally nuanced visual stories for a range of downstream tasks from education to marketing. 
 

%% file: content/10-acknowledgements.tex
\section{Acknowledgements}

We are grateful to Monojit Choudhury (Professor, MBZUAI) for his insights on CSIs and  Mohit Jain (Principal researcher, MSRI) for his valuable feedback on the paper. We also extend our gratitude to Varun Gumma (SCAI Center Fellow, MSRI) and Pranjal Chitale (Research Fellow, MSRI) for their discussions on the evaluation metric and Sanoop Menon (Research Intern, MSRI) for his support during the data analysis process. Special thanks to all the reviewers and participants who contributed their time and insights to this study. Lastly, we are thankful to the Compliance Team at Microsoft for their swift and timely support, without which this study and paper would not have been possible.

%% file: content/appendix/appendix.tex
\newpage
\clearpage

\section{LLM prompts}
\label{appendix:a-prompts}

The following section details out the prompts used for GPT-4-Turbo for each step in Kahani's pipeline (Refer Section \ref{sec:Kahani}). Each prompt consists of two parts - System Prompt and User Prompt. Variables are highlighted in curl braces (\{\}). Certain operations were broken up into two or more parts, where we daisy chained the output of one LLM call into another to improve the performance.

\subsection{Extracting Culture}
\label{appendix:b-extracting-culture}

\begin{mybox}
    \textbf{System Prompt:}
\\\\
You are an expert at local cultures and understanding cultural nuances. You are able to pick up on the stereotypes and extrapolate from it. Based on the user's input, extract as much additional cultural context, e.g. race, caste, creed, location, socio-economic background, religion, cultural beliefs. If some information is not present, do not include it. Provide only the updated culture notes. Be brief and avoid explanation. Enhance the cultural context rather than overwriting it.

current culture notes: \\
\{cultural\_context\}
\\
\\
    \textbf{User Prompt:}
\\\\
\{user\_input\}
\end{mybox}

\subsection{Writing Story}
\label{appendix:b-write-story}
\begin{mybox}
    \textbf{System Prompt:}
\\\\
You are an expert storyteller for children in the Global South. You are smart and witty. The reading level cannot exceed grade 5 and a maximum of 500 words. Update the current draft of the story. Use the culture notes to situate the story. \\
\\
Current draft: \\
\{story\}
\\\\
Culture notes: \\
\{cultural\_context\}
\\\\
\textbf{User Prompt:}
\\\\
\{user\_input\}
\end{mybox}

\subsection{Character Focus (Extract Characters)}
\label{appendix:b-extracting-characters}
\begin{mybox}
    \textbf{System Prompt:}
\\\\
You are an expert storyteller for children. Extract a maximum of 3 characters that are  critical to the story arc. Especially a character that's involved in the climax of the story. \\
\\
Use the following notes to help you: \\
\\
1. Characters must be individuals and cannot refer to a group or objects. \\
2. Focus only on the visual features of each character.  \\
3. Do not add details about their abilities or capabilities in the description.  \\
4. Include the name of the location and specific age in each character description. \\
5. Generate the age, skin color, hair color, and dress of the characters based on the location and plot of the story. \\
6. Avoid mentioning generic dress details such as "traditional dresses." Instead, be specific about the type and color of the characters' attire. \\
7. Do not add age and dress details for animals. \\
 
Think through these guidelines and then respond in valid JSON list format: \\
\\
```json
[\{"name":"", "description":"<visual description of the character from (a prominent location e.g. state or city) and details like age, skin color, hair color and hair style, the dress details>"\}]
```

A few examples for valid character descriptions: \\

[\{"name": "Geetha," "description": "A young girl from Karnataka about 7-year-old with medium brown skin and long, wavy black hair often styled in braids adorned with bright orange marigolds. She has cheerful brown eyes that sparkle with enthusiasm. Geetha prefers wearing a saffron-colored frock that complements her vibrant personality and her love for the outdoors."\}]

[\{"name": "Ramu Mama," "description": "An elderly man from Chennai, in his late 60s, wearing a light blue cotton shirt with rolled-up sleeves and white mundu. He has a fair complexion with a prominent white mustache. His hair is salt and pepper-colored, neatly combed back."\}]
\\\\
\textbf{User Prompt:}
\\\\
\{story\}
\end{mybox}

\subsection{Scene Composition}
\label{appendix:b-scene-composition}
\subsubsection{Story Scene Segmentation}
\begin{mybox}
    \textbf{System Prompt:}
\\\\
You are a top notch editor of children's illustrated books. You have an expertise in editing the story into critical scenes such that young readers find it engaging and understand the story even if they don't read the text. \\
\\
Generate 4 critical scenes for the story provided by the user to capture complete arc based on the following guidelines: \\

1. Scene 1: Introduction of the protagonist in his context. \\
2. Scene 2: Introduction of the conflict. \\
3. Scene 3: Climax of the story. \\
4. Scene 4: Conclusion on an uplifting note. \\
               
Respond in JSON List without backticks: \\
\text\{[<scene description>,<scene description>]\}
\\\\
\textbf{User Prompt:}
\\\\
Story: \\
\{story\} \\
 \\
characters: \\
\{characters\} \\
\end{mybox}

\subsubsection{Scene Planning}
\begin{mybox}
    \textbf{System Prompt:}
\\\\
You are a top-notch illustrator of children's books. You have expertise in composing the scene such that it captures the essence. You are provided the following story and characters
 \\
Story: \\
\{story\} \\
 \\
Characters: \\
\{characters\} \\

For the given scene context, generate precise outputs for the narrator and junior illustrator, providing information on the backdrop and the pose and expressions of each of the characters. \\
\\
Please note: \\
\\
1. Do not create more characters than those provided by the user.  \\
2. Scene should contain a maximum of 2 characters. \\
3. Geographical and Cultural Accuracy: If the story mentions real-world locations, ensure that each scene's backdrop accurately portrays the location's geography and iconic cultural landmarks. This will add a layer of authenticity and immersion. \\
4. Cultural Elements: When appropriate, incorporate culturally significant items, symbols, or motifs that enhance the storytelling and enrich the scene's context. \\
5. Give special emphasis to emotions and feelings. \\
6. Bring out the tension or turmoil in the protagonist. \\

Response in JSON Object format without backticks: \\
\\
\{
  "narration": "<Narrator's voice over. Short and Conversational, Grade-5 reading level in the Global South.>",
  "backdrop":"<A detailed visual description for the illustrator, highlighting the scene setting, key elements, and any specific geographical or cultural features.>", 
  characters:\{"<name of character, don't combine characters and don't add new characters>":"<pose of character>,<face expression>"\}
\}

\textbf{User Prompt:}
\\\\
scene context \\
\{context\} \\
\end{mybox}

\subsection{Creating Visuals}
\label{appendix:b-visuals}
\begin{mybox}
    \textbf{System Prompt:}
\\\\
You are an expert visual artist with a special understanding of children's book illustrations. Your task is to generate a scene description (prompt) based on the user's description of the backdrop, narration, character's actions in the current scene, and their visual description. \\
\\
Definition of user-provided information: \\
\\
Backdrop - the description of the background or scene \\
Narration - the text that will be written below this scene in the book published \\
Character action - what the characters are doing in this scene \\
Character description - detailed notes on the character visual description \\

Guidelines to generate a good prompt: \\
\\
1. Needs to be short, concise and specific.  \\
2. Needs to be closely aligned with the narration.  \\
3. Capture the action in the scene and the spatial relationship between characters. \\
4. If the character's attire uses local words like lungi, pattu pavada, replace the local word with a concise, one-line definition, emphasizing its distinctive features, cultural significance, and style elements, capturing its essence with precision and elegance.  \\
5. Make sure the prompt captures only the visually striking features of the character. \\
6. Do not use character names like Raj and Simran. Instead use a generic identifier Boy, Girl, Man, etc. \\

Prompt format: \\
<char 1 identifier and char 2 identifier>,((char1 age in dress details, doing a specific action:1.2)),((char2 age in dress details, doing a specific action:1.2)),(details of the background),(Kids illustration, Pixar style:1.2), masterpiece, sharp focus, highly detailed, cartoon

Examples: \\
Boy and dog, ((boy in a bright yellow cotton T-shirt and navy blue shorts, running excitedly, a big smile, short curly hair bouncing:1.2)), ((golden-furred dog with red collar, running with tongue out, tail wagging, alert ears:1.2)), (Marina Beach at sunrise, panoramic ocean view, Chennai cityscape, sand with colorful kites, iconic palm trees, South Indian stalls with architectural motifs), (Kids illustration, Pixar style:1.2), masterpiece, sharp focus, highly detailed, cartoon

Girl and Boy, ((Girl 10 years old, red sweater, light brown skin, straight dark hair, carrying gold chalice:1.2)), ((Boy 11 years old, navy blue jacket, curly brown hair, carrying a gold box:1.2)), Twilight, aged building with peeling paint and overgrown ivy, warm hues of the setting sun, twinkling town lights in the distance, silhouette of mountains, (Kids illustration, Pixar style:1.2), masterpiece, sharp focus, highly detailed, cartoon
\\\\
\textbf{User Prompt:}
\\\\
Scene backdrop: \\
\{backdrop\} \\
 \\
Character action \\
\{action\} \\
 \\
character description \\
\{description\} \\
 \\
Narration of Scene: \\
\{narration\} \\
\end{mybox}

\section{Sample generated stories}
\label{sec: Sample-generated-stories}

\subsection{ChatGPT4 with DALL-E3}
Once upon a sunny morning in Dalhousie, Preeti and her friends were walking through the bustling local markets, filled with the sweet aroma of flowers and spices. Their laughter mingled with the sounds of the busy shoppers around them. Suddenly, they spotted an old, forgotten building peeking out from behind the newer shops. It looked like no one had entered it for years, covered in vines and dust, with a secret entrance barely visible.

Curiosity sparked in their eyes, and they couldn't resist exploring it. As they pushed the heavy door open, a cloud of dust welcomed them. Inside, it was like stepping into another world, with shadows dancing on the walls and the air filled with whispers of the past.

As they ventured deeper, they found mysterious symbols on the walls and old, dusty books that crumbled at their touch. Suddenly, they heard a soft thud behind them. Turning around, they saw a shadow darting through the hallway. Their hearts raced, but their curiosity pushed them forward.

They followed the shadow to a hidden room, where they found a treasure chest. Just as they were about to open it, the floor beneath them gave way, and they fell into a dark tunnel. After what felt like hours of wandering in the dark, they found a light leading them back outside, emerging on the other side of the hill, with the chest still in their hands.

Opening the chest, they found it filled with ancient artifacts and a map of Dalhousie showing hidden paths and treasures. They realized that the building wasn't just old and forgotten; it was a gateway to the town's secret history.
With hearts full of joy and minds filled with memories, they promised to keep the adventure a secret, a bond that would tie them forever. And so, Preeti and her friends returned to the markets, their spirits lifted and their friendship stronger than ever, knowing they had shared an adventure they would never forget.

\begin{figure}[!h]
\centering
\includegraphics[width=\linewidth]{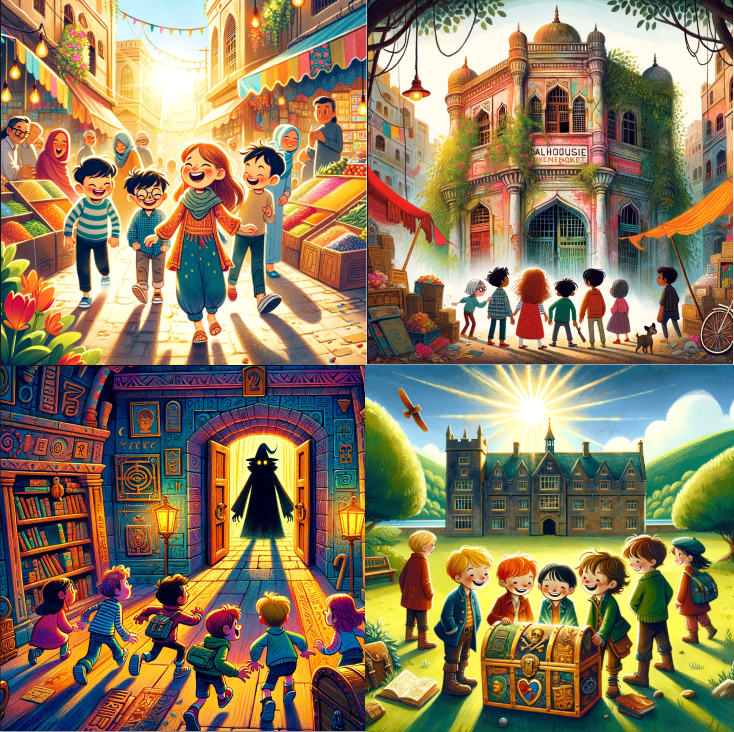}
\caption{ChatGPT + Dall-E Visuals}
\label{fig:ChatGPT4's version of Preeti}
\end{figure}

\subsection{Kahani}
\label{appendix:a-preeti-story-kahani}
In the heart of Dalhousie, where the sun plays hide and seek with the mountains, a group of friends, led by young Preeti, set out to embrace the day. Preeti, with curious eyes and a heart full of adventure, led her band of explorers through the bustling local markets. The sounds of haggling, the rich smells of street food, and the colorful displays of small family shops surrounded them.

As they wandered, they chanced upon an odd sight — an old, forgotten building that looked as though it hadn’t seen visitors in many a moon. Its walls whispered tales of ages past, and though the paint was peeling and the doors creaked, it stood with an air of mystery. “You think ghosts live there?” Preeti’s friend, Arjun, said with a mix of fear and excitement in his voice. “There’s only one way to find out,” Preeti replied, her spirit unshaken by the eerie reputation that such places often held in their community. Ignoring the hesitancy of her friends, she led the way through the hidden entrance, which was covered in a tapestry of ivy.

Inside, the air was cool, and sunlight filtered through the cracks, dancing in the dust-filled space. Cobwebs adorned every corner, and a musty smell lingered, but the intrigue was too much to resist. The friends’ footsteps echoed as they roamed, discovering trinkets and artifacts, remnants of life gone by.

Suddenly, a shadow flitted across the wall. Gasps escaped the children’s lips. “Don’t be scared,” Preeti whispered, sure it was only their imagination. “Legends and superstitions are just stories, like the ones our grandparents tell.”

Tiptoeing forward, they saw it was just a playful monkey making its home among the relics. Laughter bubbled up, breaking the tension, and they continued their journey. In the heart of the building, they found a mural. It depicted the colors and shapes of Dalhousie, with mountains and trees painted with such care that the friends felt they could almost step into the scene.

Preeti reached out her hand, tracing the lines of the artwork. As she did, a surprise awaited them. A section of the mural moved, revealing a hiding place. Inside, there were old coins, faded photographs, and letters penned with love and longing. “Our ancestors left stories here for us to find,” Preeti said, her eyes bright. The friends nodded, feeling suddenly connected to the history of Dalhousie.

As the sun dipped low in the sky, Preeti and her friends emerged from the building, their arms filled with treasures and tales to tell. They knew that some might not believe their adventure, that creeping through abandoned places was not common for kids their age. But Preeti believed that every place, no matter how old, held a story waiting to be heard. And it was these stories that made every nook and cranny of Dalhousie magical, even an old, dilapidated building that time had nearly forgotten.

As they walked away, the building seemed less forlorn, its silence now rich with newfound memories. And Preeti, with a spirit as bright as the sun above Dalhousie, already dreamed of their next adventure.

\newpage

\begin{figure}[!h]
\centering
\includegraphics[width=\linewidth]{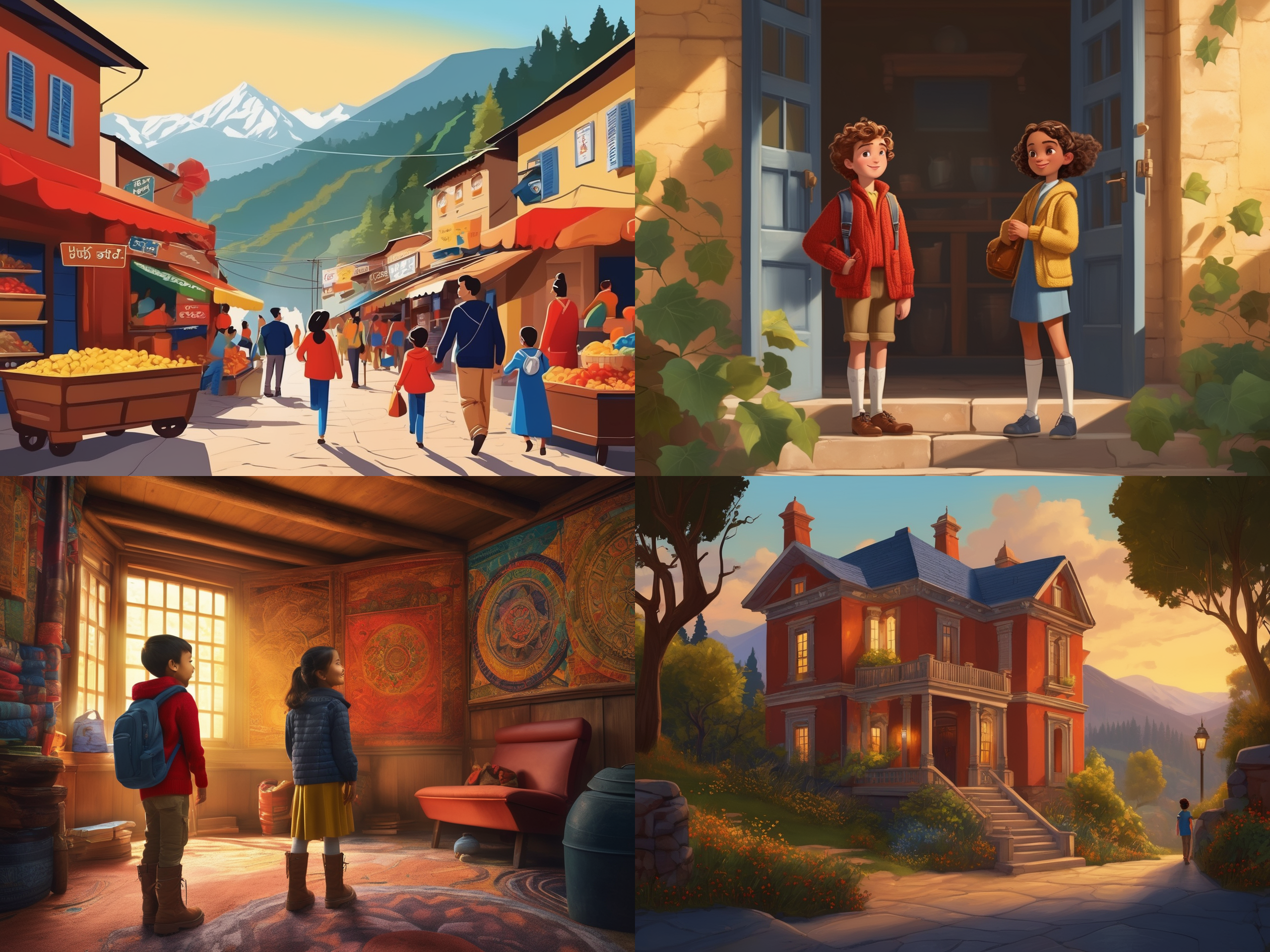}
\caption{Kahani's visuals}
\label{fig:Our pipeline's version of Preeti}
\end{figure}

\section{CSI categories and examples}
\label{appendix:csi-categories-examples}

\begin{table}[h]
\centering
\scriptsize
\begin{tabular}{@{}ll@{}}
\toprule
\textbf{CSI Category}                                & \textbf{CSI examples}                                                \\ \midrule
{\color[HTML]{3531FF} \textbf{Ecology}}              & Rice paddies, Backwaters, Coconut shell, Palm leaf                   \\
{\color[HTML]{009901} \textbf{Public life}}          & Community, Village                                                   \\
{\color[HTML]{F56B00} \textbf{Social life}}          & Local market, Small family shops, Local street food, Boarding school \\
{\color[HTML]{6200C9} \textbf{Personal life}}        & Jackfruit chips, Rice, Jalebi                                        \\
{\color[HTML]{963400} \textbf{Customs and pursuits}} & Barefeet, Legends and superstitions, Cricket match, Kabbadi          \\
{\color[HTML]{FE0000} \textbf{Private passions}}     & Mural, Reading comic book, Students Playing cards, Diwali            \\
{\color[HTML]{F004E5} \textbf{Proper Nouns}}         & Karuthamma, Uncle Rajan, Mr. Kumar, Dehradun, Bengal, Gujarat        \\ \bottomrule
\\
\end{tabular}%
\caption{CSI categories and examples for each category from the stories generated by Kahani}
\label{table:csi-categories-examples}
\end{table}

\section{User Study}

\subsection{Participant Task Instructions}

\begin{figure*}[ht]
\centering
\begin{tikzpicture} 
\definecolor{lightred}{RGB}{255,191,191}
\definecolor{darkred}{RGB}{191,0,0}
\definecolor{lightblue}{RGB}{191,191,255}
\definecolor{darkblue}{RGB}{0,0,191}
\definecolor{lightgreen}{RGB}{191,255,191}
\definecolor{darkgreen}{RGB}{0,191,0}
\definecolor{lightyellow}{RGB}{255,255,191}
\definecolor{darkyellow}{RGB}{191,191,0}
\definecolor{lightgrey}{RGB}{211,211,211}
\definecolor{darkgrey}{RGB}{128,128,128}
\node[draw, rectangle, minimum width=10cm, minimum height=3cm, fill=lightgrey, draw=darkgrey, rounded corners=5pt, inner xsep=15pt, inner ysep=15pt] 
{\begin{minipage}{15cm}

\textbf{Introduction} \\ \\
In this task, you will identify the cultural aspects in both text and visual stories generated by two AI tools. Your assessment will help us understand how well the stories capture cultural nuances. \\
 
\textbf{Task Overview} \\
You will be provided with two pairs of stories and their accompanying visuals. For each pair, follow these steps:

\begin{enumerate}
    \item \textbf{Read the Story}
\begin{itemize}
        \item Carefully read through the story text. 
        \item Pay close attention to cultural elements such as names, traditions, geography, food, settings etc.
        \item As you read, highlight specific cultural elements (both correct and incorrect) that stand out.
        \item For each highlighted element, provide a brief comment explaining your reasoning. You can either type your comments or, if you prefer, give verbal feedback.
\end{itemize}
    \item \textbf{Examine the Visuals:} 
\begin{itemize}
        \item After reading the story, closely examine the corresponding visuals.
        \item Highlight the cultural elements in the visuals, such as characters, environments, artifacts etc., that align with or diverge from the cultural context of the story.
        \item Highlight the cultural elements in the visuals, and provide a brief comment explaining your reasoning, just like you did with the text.
\end{itemize}
    \item \textbf{Rate the Story and Visuals:}
\begin{itemize}
        \item After reviewing each story and its visuals, assign a rating based on the given metrics (on the screen).
        \item Provide comments explaining why you assigned a particular rating, noting specific strengths or weaknesses in how the story and visuals represented cultural aspects.
         \item You may type your comments or give verbal feedback if that is more comfortable.
\end{itemize}
\end{enumerate}
 
\textbf{Additional notes}
\begin{itemize}
    \item If you encounter any part of the story or visuals that feels offensive, irrelevant, or clearly off-topic, please flag it for review by typing a comment or verbally specifying the issue.
    \item Take your time to evaluate each story and visual pair carefully. Your thoughtful feedback is crucial for the success of this study.
    \item Below are some shortcuts to help you navigate the FigJam tool:
    \begin{itemize}
        \item Scroll to zoom in or out
        \item Press \textbf{H} for the hand tool to move around the board
        \item Press \textbf{M} to use the marker tool and change colors
        \item Press \textbf{S} for sticky notes to leave comments
    \end{itemize}
\end{itemize}
 
\end{minipage}};
\end{tikzpicture}
\caption{Detailed task instructions provided to the participants}
\label{fig:task_instructions}
\end{figure*}

\newpage
\subsection{Semi-structured interview questions}

\begin{figure*}[ht]
\centering
\begin{tikzpicture} 
\definecolor{lightred}{RGB}{255,191,191}
\definecolor{darkred}{RGB}{191,0,0}
\definecolor{lightblue}{RGB}{191,191,255}
\definecolor{darkblue}{RGB}{0,0,191}
\definecolor{lightgreen}{RGB}{191,255,191}
\definecolor{darkgreen}{RGB}{0,191,0}
\definecolor{lightyellow}{RGB}{255,255,191}
\definecolor{darkyellow}{RGB}{191,191,0}
\definecolor{lightgrey}{RGB}{211,211,211}
\definecolor{darkgrey}{RGB}{128,128,128}
\node[draw, rectangle, minimum width=10cm, minimum height=3cm, fill=lightgrey, draw=darkgrey, rounded corners=5pt, inner xsep=15pt, inner ysep=15pt] 
{\begin{minipage}{15cm}

\textbf{Story Script}
\begin{enumerate}
    \item Which of the stories do you like? Why?
    \item Which one is the most culturally nuanced? What aspects make it so?
    \item How well do you think the story’s setting and characters represent real-life scenarios or cultures? Can you provide examples?
\end{enumerate}

\textbf{Scene visuals}
\begin{enumerate}
    \item What are your thoughts on how culture is represented in these images/visuals?  
    \item How well do the visuals complement the story?Is there a particular detail that stands out to you?
    \item Is there anything missing from the visuals that you think should have been included to better tell the story or represent its cultural aspects?
    \item Do you find the story/visuals relatable? If so, how? 
\end{enumerate} 

\textbf{General}
\begin{enumerate}
    \item Have you explored any generative AI tools for creating text-based or image-based stories?
    \item Which software/tools have you used before?  
    \item Can you share specific use cases where you have used these tools?  
    \item Do you have any concerns or reservations about using AI for storytelling
\end{enumerate}

\end{minipage}};
\end{tikzpicture}
\caption{Semi-structured interview questions}
\label{fig:interview questions}
\end{figure*}